\documentclass{article}
\usepackage{iclr2026_conference,times}

\usepackage{amsmath,amsfonts,bm}

\def\eqref#1{equation~\ref{#1}}

\def\1{\bm{1}}

\DeclareMathAlphabet{\mathsfit}{\encodingdefault}{\sfdefault}{m}{sl}
\SetMathAlphabet{\mathsfit}{bold}{\encodingdefault}{\sfdefault}{bx}{n}

\usepackage{amsmath}
\usepackage{amssymb}
\usepackage{booktabs}
\usepackage{graphicx}
\usepackage{array}
\usepackage{hyperref}
\usepackage{url}
\usepackage{multirow}
\usepackage{enumitem}
\usepackage[ruled,lined]{algorithm2e}
\usepackage[table]{xcolor} 
\usepackage{tabularx}      

\definecolor{headergray}{rgb}{0.93, 0.95, 0.97}     
\definecolor{highlightrow}{rgb}{0.94, 0.96, 0.97}   
\newcommand{\best}[1]{\textcolor{red}{\textbf{#1}}}
\newcommand{\second}[1]{\textcolor{blue}{\underline{#1}}}

\newcolumntype{C}{>{\centering\arraybackslash}X}

\title{Latent Reward Registers for \\ Diffusion Preference Alignment}

\author{
  Zipeng Feng$^{1}$\thanks{Equal contribution.}, Yuanshen Guan$^{2}$\footnotemark[1], Chengru Song$^{1}$, Zhiwei Xiong$^{2}$\thanks{Corresponding author.}, Peiqin Sun$^{1}$\footnotemark[2] \\
  $^1$Kling Team \quad $^2$University of Science and Technology of China \\
  \texttt{guanys@mail.ustc.edu.cn} \quad \texttt{zwxiong@ustc.edu.cn} \\
  \texttt{\{fengzipeng, songchengru, sunpeiqin\}@kuaishou.com} 
}

\newcommand{\method}{Latent Reward Registers}
\newcommand{\samplingmethod}{Reward-Guided Sampling}
\newcommand{\guidance}{\alpha}
\newcommand{\rms}{\operatorname{RMS}}
\newcommand{\unitrms}{\operatorname{unitRMS}}

\iclrfinalcopy

\begin{document}

\maketitle
\lhead{}

\begin{abstract}
Aligning diffusion models with human preferences usually relies on a sparse terminal reward evaluated on the final generated samples, which creates a severe temporal credit-assignment problem across the denoising process. We propose \emph{Latent Reward Registers}, a mechanism that estimates terminal preference directly from intermediate noisy latents. Learnable, position-free register tokens are appended as an auxiliary read path to a frozen Diffusion Transformer (DiT), extracting preference signals without altering the generator's hidden states or velocity field. The resulting dense, differentiable reward field spans the full denoising trajectory and supports two alignment strategies. For training, \emph{Reward-Gradient On-Policy Distillation} (RG-OPD) converts this dense reward field into per-step targets at states visited by the current generator, replacing rollout-intensive policy gradients with direct on-policy distillation. For inference, \emph{Reward-Guided Sampling} (RGS) steers trajectories with magnitude-matched reward-gradient corrections and no parameter updates. Empirically, at high noise levels ($t{=}0.8$) the registers reach the highest pairwise accuracy among the evaluated latent reward models. RG-OPD outperforms online reinforcement learning baselines while reducing GPU hours by up to $33\times$. RGS achieves significant reward improvement with a favorable reward--quality balance against training-free baselines.
\end{abstract}

\section{Introduction}

Aligning diffusion and flow-matching models with human preferences is most commonly formulated as reward maximization: an external reward model assigns a scalar score to the final generated image, and the generator is trained to maximize this score \citep{refl,draft,AlignProp,black2023training,liu2025flowgrpo,zheng2025diffusionnft}. Regardless of the specific reward optimization method, they share a fundamental bottleneck: the reward is evaluated only on the fully generated image at the end of the denoising trajectory.

Existing approaches differ primarily in how they optimize towards the terminal reward. \emph{Reward backpropagation} \citep{refl,wu2024drtune,draft,AlignProp} differentiates the reward through the denoising chain. This provides direct, low-variance gradients, but requires storing the full trajectory in memory and is vulnerable to reward over-optimization, in which the generator exploits the reward model at the expense of perceptual quality and diversity \citep{draft,AlignProp}. \emph{Online reinforcement learning (RL)} \citep{black2023training,liu2025flowgrpo,xue2025dancegrpo,zheng2025diffusionnft} instead formulates denoising as a sequential policy and optimizes it with rollout-level policy gradients. By optimizing against sampled outcomes rather than differentiated rewards, this approach preserves diversity and mitigates over-optimization. However, relying on a single terminal scalar introduces high-variance gradients and poor sample efficiency.

Both paradigms inherently suffer from a \emph{temporal credit-assignment} problem. Because the delayed terminal reward cannot be attributed to individual denoising steps, the training objective fails to identify which trajectory decisions drive the final improvement. Furthermore, lacking a step-wise signal, these methods cannot directly steer the sampling process at test time.

We argue that preference reward should instead be extracted where it is generated, along the denoising trajectory rather than at its endpoint. Our key insight is that the internal representations of a Diffusion Transformer (DiT) naturally encode early evidence of the trajectory quality. Even under high noise levels, intermediate hidden states reflect the emerging global layout, prompt alignment, and visual structure. If the expected terminal reward can be read directly from any intermediate noisy latent, sparse endpoint feedback can be converted into a dense, step-wise reward signal, thereby bypassing the credit-assignment bottleneck.

Building on this insight, we introduce \emph{Latent Reward Registers}, a lightweight mechanism that converts a frozen DiT into a step-wise reward model (Figure~\ref{fig:teaser}). We prepend a small set of learnable, position-free register tokens to the noisy input sequence. These registers are trained to aggregate reward-relevant evidence from intermediate representations. A readout head then maps the aggregated evidence to the expected terminal preference at any given denoising step. Crucially, because this readout is differentiable with respect to the latent state, it provides a local reward gradient at negligible computational cost, explicitly indicating how intermediate latents should be perturbed to maximize the expected terminal preference.

This dense, differentiable reward signal supports alignment at both training and inference time. For training, \emph{Reward-Gradient On-Policy Distillation} (RG-OPD) converts the register gradients into per-step optimization targets along the generator's own rollouts, replacing rollout-level policy gradients with local, low-variance supervision. For inference, \emph{Reward-Guided Sampling} (RGS) applies magnitude-matched reward-gradient corrections during sampling, requiring neither parameter updates nor image decoding, and allowing multiple reward objectives to be composed at test time.

Experiments on SD3-Medium and FLUX.1-dev show that the registers predict terminal preference accurately even at high noise levels, attaining the highest pairwise accuracy among the evaluated latent reward models (Section~\ref{sec:exp-accuracy}). Building on this signal, RG-OPD surpasses online RL baselines in alignment quality while reducing GPU hours by up to $33\times$ (Figure~\ref{fig:training-efficiency}). Furthermore, RGS significantly outperforms existing training-free sampling methods, achieving a superior reward--quality trade-off (Section~\ref{sec:exp-guidance}). Our contributions are summarized as follows:

\begin{figure}[t]
    \centering
    \includegraphics[width=\linewidth]{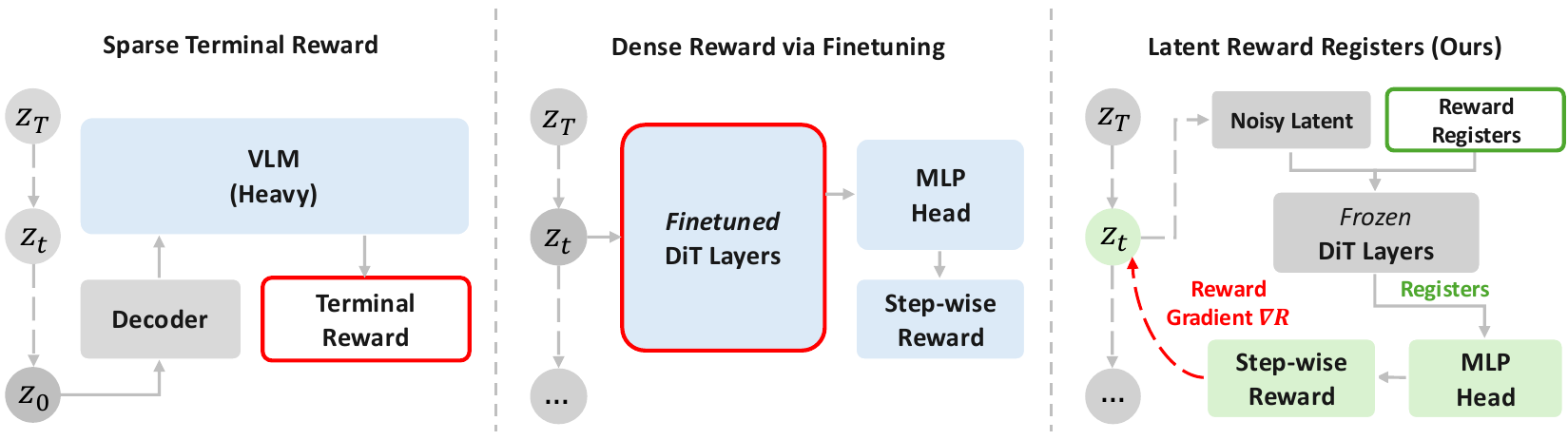}
    \vspace{-0.5cm}
    \caption{\textbf{Latent Reward Registers vs. existing paradigms.} Unlike conventional alignment methods that rely on delayed, sparse feedback from a clean image (Left) or require intrusive fine-tuning of the generator (Middle), our approach (Right) prepends learnable registers to a frozen DiT. This enables the direct estimation of step-wise rewards from intermediate noisy latents ($z_t$). Consequently, we derive dense, differentiable reward gradients ($\nabla R$) for generation while keeping the pretrained generative dynamics fully intact.}
    \vspace{-0.6cm}
    \label{fig:teaser}
\end{figure}

\begin{itemize}[nosep, leftmargin=1em]
    \item \textbf{Latent reward readout from frozen generators.} We propose Latent Reward Registers, which estimate the expected terminal preference directly from intermediate noisy latents of a frozen DiT. At high noise levels, the registers attain the highest pairwise accuracy among latent reward models.
    \item \textbf{Training-time alignment from dense reward signals.} We show that register-derived reward gradients can replace rollout-level policy gradients: RG-OPD surpasses online RL baselines in alignment quality on SD3-Medium and FLUX.1-dev while reducing GPU hours by up to $33\times$.
    \item \textbf{Inference-time alignment from the same signal.} We demonstrate that the register gradients also steer sampling directly: RGS improves target rewards without parameter updates and achieves a better reward--quality trade-off than existing training-free sampling methods.
\end{itemize}


\section{Related Work}
\label{sec:related-work}

\paragraph{Diffusion Preference Alignment.}
Preference alignment for text-to-image generators can be applied at training time or inference time. Training-time methods can be categorized into three families: reward backpropagation differentiates through the denoising chain to optimize a reward model \citep{refl, draft, AlignProp}, online reinforcement learning treats the denoising process as a policy and optimizes it with policy gradients \citep{liu2025flowgrpo, zheng2025diffusionnft}, and direct preference optimization (DPO) trains the generator on pairwise preferences without explicit reward modeling \citep{wallace2023diffusiondpo}. All three methods derive reward only from the final clean sample, and each pays a different cost. Reward backpropagation provides direct optimization signal, but its gradients must pass through the full denoising chain and fade before reaching the early steps that decide global structure and semantics. Online RL mitigates reward hacking by exploring the model's own distribution, but involves high variance and poor sample efficiency. The offline objective of DPO becomes biased once the generator drifts from the original policy, and its terminal reward is broadcast uniformly across denoising steps. Inference-time alignment avoids retraining, but existing schemes are computationally expensive, relying either on iterative search and optimization \citep{yeh2025demon, tang2025dno} or on guidance gradients that become unstable at high noise levels \citep{song2023lgd}. Our method instead extracts a dense, differentiable, step-wise reward in the latent space, making alignment efficient at both training and inference time.

\vspace{-0.2cm}
\paragraph{Latent Reward Models.}
Most reward models operate on clean, decoded images \citep{xu2023imagereward,kirstain2023pickapic,ma2025hpsv3}. Applying them to intermediate states would require repeated expensive decoding, and the resulting signals are unreliable on heavily corrupted inputs. Recent latent reward models instead evaluate noisy states directly, drawing evidence from selected internal features \citep{zhang2025lrm}, early cross-attention maps \citep{cui2026diffusionprobegeneratedimage}, separate process reward models \citep{mi2025videopavrm,go2026stitchvm}, or auxiliary heads that predict step-wise rewards \citep{yang2024dense}. Diffusion-native variants go further, training noise-aware scorers inside the generator itself \citep{liu2026dinalrm}. Existing methods tend to rely on hand-picked features or additional value networks with substantial compute. Latent Reward Registers instead prepend lightweight, position-free register tokens to a frozen DiT, building a differentiable reward interface without hand-crafted feature heuristics and without altering the base generator, which directly supports alignment in both training and inference.

\section{Method}
\label{sec:method}

\begin{figure}[t]
\centering
\includegraphics[width=1\linewidth]{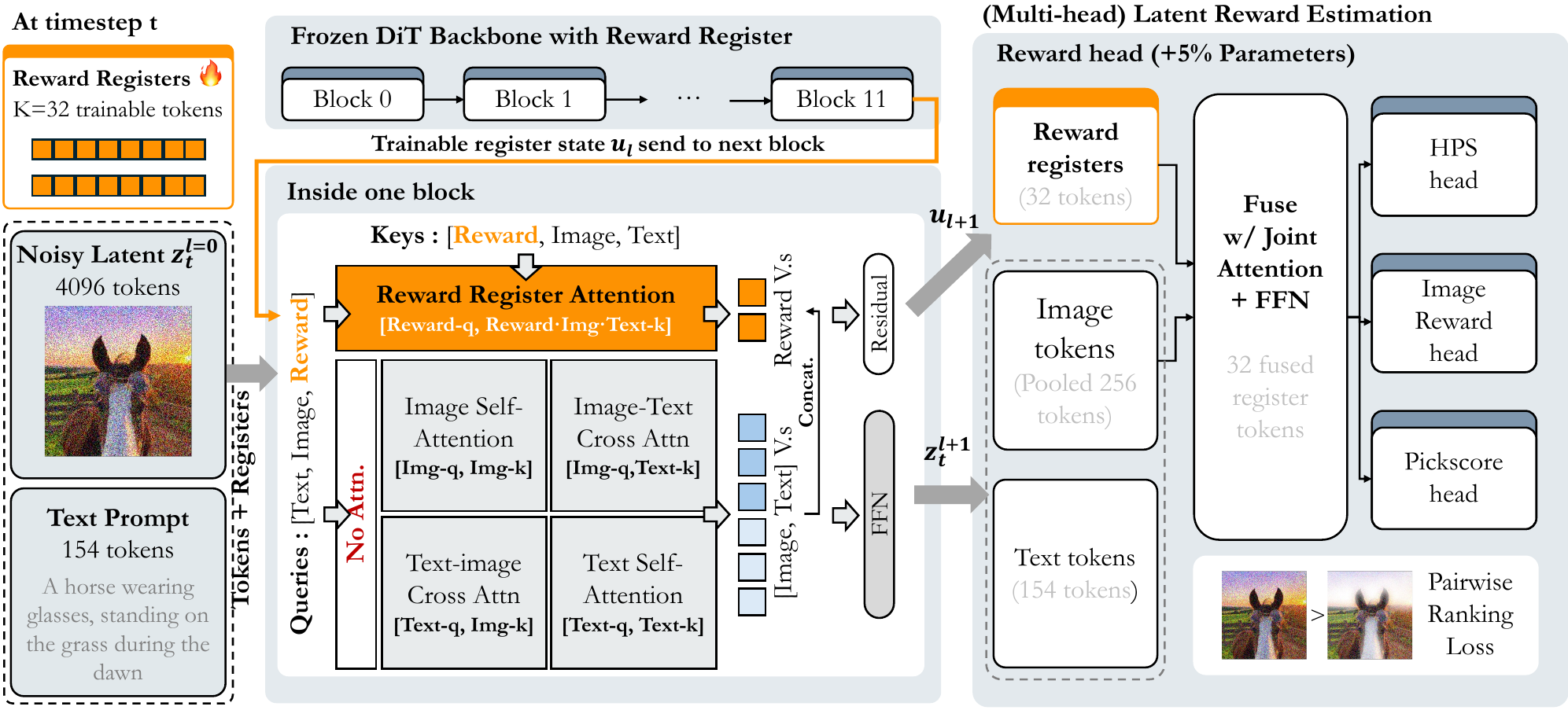}
\vspace{-0.5cm}
\caption{\textbf{Overview of the \method\ mechanism.} Learnable registers form a non-intrusive side stream through a frozen DiT. The registers read frozen image and text features through an auxiliary read path, thereby preserving the original velocity field. A lightweight fusion module combines the final register states with intermediate backbone features for multi-head latent reward estimation under pairwise ranking supervision.}
\vspace{-0.5cm}
\label{fig:reward-register}
\end{figure}

To address the temporal credit-assignment problem in preference alignment, we construct a dense, differentiable reward signal directly from intermediate latent states, providing a unified supervision mechanism for both training and inference time alignment. The proposed framework consists of three core components:

\begin{itemize}[nosep, leftmargin=1em]
\item \textbf{Latent Reward Registers} (Sec.~\ref{sec:reward-register}): A lightweight mechanism that predicts the expected terminal preference using learnable register tokens and frozen DiT features. This auxiliary stream extracts step-wise reward signals from intermediate noisy latents without perturbing the original velocity field (Fig.~\ref{fig:reward-register}).
\item \textbf{Reward-Gradient On-Policy Distillation} (Sec.~\ref{sec:rg-opd}, Algorithm~\ref{alg:rg-opd}): A training-time alignment that constructs detached step-wise optimization targets along the student sampling trajectory. This bypasses backpropagation through complete denoising chains and eliminates the high variance of rollout-level policy gradients.
\item \textbf{Reward-Guided Sampling} (Sec.~\ref{sec:reward-guided-sampling}, Algorithm~\ref{alg:reward-guided-sampling}): An inference-time alignment that steers latent trajectories using magnitude-matched, step-wise reward gradients, requiring no additional parameter updates.
\end{itemize}

\subsection{Preliminaries and Notation}
\label{sec:preliminaries}

We work in the latent space of a frozen VAE \citep{rombach2022latent}, where an image $x$ is encoded as $z_0=\operatorname{Enc}(x)$ and reconstructed as $x=\operatorname{Dec}(z_0)$. Let $c$ denote a text prompt and $z_t\in\mathbb{R}^{D}$ the latent state at timestep $t\in[0,1]$, with $t=1$ denoting noise and $t=0$ data. The probability-flow ODE $\mathrm d z_t/\mathrm d t=v(z_t,t,c)$ is integrated from noise to data \citep{lipman2023flow}. We distinguish the frozen pretrained field $v_{\mathrm{ref}}$ from the trainable student field $v_\theta$.

Let $\{\tau_i\}_{i=0}^{N}$ be a decreasing timestep schedule with $\tau_0=1$, $\tau_N=0$, and $\Delta\tau_i:=\tau_{i+1}-\tau_i<0$. Under the Euler discretization used by our primary method, the student and reference fields induce the deterministic one-step transitions
\begin{equation}
\begin{aligned}
    \mu_{\theta,i}(z_{\tau_i},c)&:=z_{\tau_i}+\Delta\tau_i\,v_\theta(z_{\tau_i},\tau_i,c),\\
    \mu_{\mathrm{ref},i}(z_{\tau_i},c)&:=z_{\tau_i}+\Delta\tau_i\,v_{\mathrm{ref}}(z_{\tau_i},\tau_i,c),
\end{aligned}
\label{eq:one-step-transitions}
\end{equation}
where native text guidance is included in each velocity prediction. Starting from $z_{\tau_0}=z_1\sim\mathcal N(0,I_D)$, the student rollout applies $z_{\tau_{i+1}}=\mu_{\theta,i}(z_{\tau_i},c)$.

Reward-model training instead uses forward-corrupted data latents
\begin{equation}
    z_t^{(a)}=(1-t)z_0^{(a)}+t\epsilon,
    \qquad \epsilon\sim\mathcal N(0,I_D),
\label{eq:forward-corruption}
\end{equation}
where $z_0^{(a)}$ is the data latent of image $a$ within a preference pair $(a,b)$. The two images share the same $t$ and $\epsilon$, ensuring an equal corruption level across the pair.

\subsection{Reward Register Mechanism}
\label{sec:reward-register}

\paragraph{Motivation.}
Standard reward models supervise only the terminal state, and applying them to intermediate states would require repeated decoding that exposes the reward signal to decoder artifacts at high noise levels. DiT representations, in contrast, encode prompt alignment, layout, and appearance early in the generation process. We therefore introduce a lightweight non-intrusive mechanism that predicts terminal preference directly from a noisy latent $z_t$. Unlike a static feature probe, the register state persists across multiple frozen DiT blocks, accumulating evidence across depth. Because it reads from the backbone without writing to the token sequence, the side stream provides differentiable step-wise supervision while preserving the original velocity field.

\paragraph{Non-intrusive reward registers.}
We initialize $K$ learnable, position-free register tokens $Q^{(0)}\in\mathbb{R}^{K\times d_h}$ as global readouts, where $d_h$ is the hidden width of the frozen DiT. For each latent state, we set $Q_t^{(0)}=Q^{(0)}$. Let $H_t^{(\ell)}=[H_{t,\mathrm{img}}^{(\ell)},H_{t,\mathrm{txt}}^{(\ell)}]$ denote the native image and text hidden states at block $\ell$ for latent $z_t$. For the initial $L$ transformer blocks, the registers share a single trainable query projection $W_Q^\phi$ while reusing the frozen key, value, output, normalization, and time-conditioning modules from each block. Letting $\bar{Q}_t^{(\ell)}$, $\bar{H}_{t,\mathrm{img}}^{(\ell)}$, and $\bar{H}_{t,\mathrm{txt}}^{(\ell)}$ denote the $t$-conditioned normalized states, the update is:
\begin{equation}
\begin{aligned}
    \mathbf{K}_t^{(\ell)} &= \left[
        \bar{Q}_t^{(\ell)}W_K^{(\ell)},\,
        \bar{H}_{t,\mathrm{img}}^{(\ell)}W_K^{(\ell)},\,
        \bar{H}_{t,\mathrm{txt}}^{(\ell)}W_{K,\mathrm{txt}}^{(\ell)}
    \right],\\
    \mathbf{V}_t^{(\ell)} &= \left[
        \bar{Q}_t^{(\ell)}W_V^{(\ell)},\,
        \bar{H}_{t,\mathrm{img}}^{(\ell)}W_V^{(\ell)},\,
        \bar{H}_{t,\mathrm{txt}}^{(\ell)}W_{V,\mathrm{txt}}^{(\ell)}
    \right],\\
    Q_t^{(\ell+1)}
    &= Q_t^{(\ell)} + \gamma_t^{(\ell)} \odot
    \operatorname{Attn}\!\left(
        \bar{Q}_t^{(\ell)}W_Q^\phi,\,
        \mathbf{K}_t^{(\ell)},\,
        \mathbf{V}_t^{(\ell)}
    \right)W_O^{(\ell)},
\end{aligned}
    \label{eq:register-update}
\end{equation}
where $[\,\cdot\,,\,\cdot\,]$ denotes concatenation along the token axis with tokens arranged as rows, $\operatorname{Attn}(\mathbf{Q},\mathbf{K},\mathbf{V})=\operatorname{softmax}(\mathbf{Q}\mathbf{K}^{\top}\!/\sqrt{d_h})\mathbf{V}$ uses bold operands for generic attention matrices to distinguish them from the register count $K$ and register states $Q$, $\gamma_t^{(\ell)}$ is the frozen backbone's $t$-conditioned attention gate, and $\odot$ denotes element-wise multiplication. This design isolates the side stream: native image and text streams are computed independently of the registers, and for single-stream architectures (e.g., FLUX.1-dev) the native sequence serves directly as the combined input context. The registers also bypass the block feed-forward networks, so the hidden states and velocity predictions of the original DiT backbone stay strictly invariant. Finally, the register states are fused with pooled frozen features from a selected block set $\mathcal{B}$ through a lightweight attention module, producing shared per-register features $U_t\in\mathbb{R}^{K\times d_h}$ for the reward heads.

\paragraph{Reward prediction and learning.}
For reward objective $m$, the latent score is
\begin{equation}
    r_\phi^{(m)}(z_t,c,t)
    = \frac{1}{K}\sum_{k=1}^{K}h_\phi^{(m)}\!\left(U_{t,k}\right),
    \qquad m=1,\ldots,M,
    \label{eq:learning-target}
\end{equation}
where $h_\phi^{(m)}$ is a reward-specific two-layer MLP. Each head is trained with the noise-aware Thurstone pairwise-ranking objective \citep{thurstone1927law} and the DiNa-LRM variance adjustment \citep{liu2026dinalrm}. The loss penalizes disagreement with the ranking induced by the corresponding endpoint reward model, while the assumed comparison variance increases with the sampled noise level $t$. Images within each pair share the same $t$ and Gaussian noise realization, ensuring an equal corruption level across the pair. The backbone stays frozen. Only the registers, shared query projection, fusion readout, and reward heads are optimized, and an exponential moving average of these parameters is used for evaluation and downstream guidance. The learned score preserves the endpoint reward ranking rather than estimating a calibrated reward value.

\paragraph{Shared reward direction.}
For objective $m$, the latent reward gradient at a trajectory state is
\begin{equation}
    g_i^{(m)}:=\nabla_{z_{\tau_i}}r_\phi^{(m)}(z_{\tau_i},c,\tau_i).
\label{eq:reward-gradient}
\end{equation}
We write $\rms(u):=\sqrt{\lVert u\rVert_2^2/D}$ for the root-mean-square of a latent tensor $u$ and $\unitrms(u):=u/\max\{\rms(u),\delta\}$ for its normalization to unit RMS, with a small numerical floor $\delta>0$. Given selected reward heads $\mathcal M\subseteq\{1,\ldots,M\}$, we combine their directions as
\begin{equation}
    g_i^{\mathcal M}:=\sum_{m\in\mathcal M}\unitrms\!\left(g_i^{(m)}\right),
    \qquad
    \widehat g_i^{\mathcal M}:=\unitrms\!\left(g_i^{\mathcal M}\right).
\label{eq:combined-reward-direction}
\end{equation}
Thus each head contributes equally before the aggregate direction is normalized. Single-objective guidance uses $\mathcal M=\{m\}$. We stop gradients through $\widehat g_i^{\mathcal M}$ when it is used to construct an RG-OPD training target. RGS is an inference-time procedure and involves no parameter gradients.

\begin{algorithm}[t]
\caption{\textbf{Reward-Gradient On-Policy Distillation (RG-OPD).}}
\label{alg:rg-opd}
\KwIn{prompt distribution $\mathcal{C}$; frozen reference field $v_{\mathrm{ref}}$; frozen reward register $\{r_\phi^{(m)}\}$; schedule $\{\tau_i\}_{i=0}^{N}$; guidance strengths $\{\guidance_i\}_{i=0}^{N-1}$; selected heads $\mathcal{M}$; optimized steps $\mathcal{J}$}
\KwOut{reward-aligned student field $v_\theta$}
Initialize $v_\theta\leftarrow v_{\mathrm{ref}}$\;
\For{each training round}{
    Sample prompts $c\sim\mathcal{C}$\;
    Sample $z_{\tau_0}=z_1\sim\mathcal N(0,I_D)$ and roll out the current student \tcp*[r]{no\_grad}
    \For{$i\in\mathcal{J}$}{
        Evaluate $\mu_{\mathrm{ref},i}=\mu_{\mathrm{ref},i}(\operatorname{sg}(z_{\tau_i}),c)$ and $\Delta_{\mathrm{ref},i}=\mu_{\mathrm{ref},i}-\operatorname{sg}(z_{\tau_i})$\;
        \eIf{$\guidance_i>0$}{
            Compute and detach $\widehat g_i^{\mathcal{M}}$ using Eq.~\ref{eq:combined-reward-direction}\;
        }{
            Set $\widehat g_i^{\mathcal{M}}\leftarrow 0$\;
        }
        Construct and store $\operatorname{sg}(\mu_{\mathrm{tar},i})$ using Eq.~\ref{eq:rg-opd-teacher}\;
    }
    \For{$i\in\mathcal{J}$}{
        Evaluate $\mu_{\theta,i}(\operatorname{sg}(z_{\tau_i}),c)$ and accumulate the loss in Eq.~\ref{eq:rg-opd-loss}\;
    }
    Perform a single optimizer step on the accumulated loss, updating only $\theta$\;
}
\end{algorithm}


\subsection{Reward-Gradient On-Policy Distillation}
\label{sec:rg-opd}

Reward-Gradient On-Policy Distillation (RG-OPD) generates step-wise supervision at states actively visited by the {current student model}, adapting the training distribution to the evolving student rather than relying on fixed offline teacher trajectories. This follows the on-policy distillation paradigm, in which teacher signals are constructed and consumed along student rollouts \citep{li2026diffusionopd, yuan2026visionopd, jiang2026dopsd}.

Each one-step operator in Eq.~\ref{eq:one-step-transitions} is a function of an arbitrary state at $\tau_i$. In particular, the reference transition is evaluated at states visited by the student, not at states produced by the reference field itself. For a student-rollout state $z_{\tau_i}$, the frozen reference field produces
\begin{equation}
    \mu_{\mathrm{ref},i}(z_{\tau_i},c)
    =z_{\tau_i}+\Delta\tau_i\,v_{\mathrm{ref}}(z_{\tau_i},\tau_i,c),
    \qquad
    \Delta_{\mathrm{ref},i}(z_{\tau_i},c)
    :=\mu_{\mathrm{ref},i}(z_{\tau_i},c)-z_{\tau_i}.
    \label{eq:rg-opd-reference}
\end{equation}
Let $\guidance_i\ge0$ be the reward-guidance strength. The reward-tilted one-step target is
\begin{equation}
    \mu_{\mathrm{tar},i}(z_{\tau_i},c)
    := \mu_{\mathrm{ref},i}(z_{\tau_i},c)
    + \guidance_i\,
    \rms\!\left(\Delta_{\mathrm{ref},i}(z_{\tau_i},c)\right)
    \widehat g_i^{\mathcal M}.
    \label{eq:rg-opd-teacher}
\end{equation}
The RMS factor expresses the correction relative to the underlying solver displacement. When $\guidance_i=0$, the target reduces to the frozen reference transition and acts as an explicit retention target. Importantly, $\mu_{\mathrm{tar},i}$ is constructed supervision, not the output of a separately parameterized teacher network.

Let $\operatorname{sg}(\cdot)$ denote stop-gradient, $p_\theta(z_{\tau_0},\ldots,z_{\tau_N}\mid c)$ the trajectory distribution induced by the current student, and $\mathcal{J}\subseteq\{0,\ldots,N-1\}$ the optimized denoising steps. The RG-OPD objective is
\begin{equation}
\begin{aligned}
    \mathcal{L}_{\mathrm{RG\text{-}OPD}}(\theta)
    &=\mathbb E_{c\sim\mathcal C}
    \mathbb E_{(z_{\tau_0},\ldots,z_{\tau_N})\sim p_\theta(\cdot\mid c)}
    \Bigg[\frac{1}{|\mathcal J|D}\sum_{i\in\mathcal J}
    \Big\|\mu_{\theta,i}\!\left(\operatorname{sg}(z_{\tau_i}),c\right) - \operatorname{sg}\!\left(\mu_{\mathrm{tar},i}(z_{\tau_i},c)\right)\Big\|_2^2\Bigg].
\end{aligned}
    \label{eq:rg-opd-loss}
\end{equation}
The outer expectation makes the on-policy state distribution explicit. During differentiation, however, sampled states and constructed targets are constants, so Eq.~\ref{eq:rg-opd-loss} is a local semi-gradient update rather than a gradient propagated through $p_\theta$. The reference generator and reward register remain frozen, and gradients update only $\theta$. Algorithm~\ref{alg:rg-opd} summarizes the procedure. Backbone-specific parameterizations and $\mathcal{J}$ are reported in Appendix~\ref{sec:appendix-settings-opd}.

\begin{algorithm}[t]
\caption{\textbf{Reward-Guided Sampling.}}
\label{alg:reward-guided-sampling}
\KwIn{prompt $c$; frozen reference field $v_{\mathrm{ref}}$; frozen reward register $\{r_\phi^{(m)}\}$; schedule $\{\tau_i\}_{i=0}^{N}$; guidance strengths $\{\guidance_i\}_{i=0}^{N-1}$; selected heads $\mathcal{M}$}
\KwOut{generated image $x$}
Sample $z_{\tau_0}=z_1\sim\mathcal N(0,I_D)$\;
\For{$i\leftarrow 0$ \KwTo $N-1$}{
    Evaluate $\mu_{\mathrm{ref},i}=\mu_{\mathrm{ref},i}(z_{\tau_i},c)$ and $\Delta_{\mathrm{ref},i}=\mu_{\mathrm{ref},i}-z_{\tau_i}$\;
    \eIf{$\guidance_i>0$}{
        Compute $\widehat g_i^{\mathcal{M}}$ using Eq.~\ref{eq:combined-reward-direction}\;
        Set $z_{\tau_{i+1}}$ using the magnitude-matched rule \tcp*[r]{Eq.~\ref{eq:guided-update}}
    }{
        Set $z_{\tau_{i+1}}\leftarrow\mu_{\mathrm{ref},i}$\;
    }
}
\Return{$x=\operatorname{Dec}(z_{\tau_N})$}\;
\end{algorithm}

\subsection{Reward-Guided Sampling}
\label{sec:reward-guided-sampling}

Reward-Guided Sampling (RGS) applies the same reward-tilted transition directly at inference time. The reference generator and reward register remain frozen, and no student model is introduced. The reference transition $\mu_{\mathrm{ref},i}$ and its displacement $\Delta_{\mathrm{ref},i}$ follow Eq.~\ref{eq:rg-opd-reference}. For an active step, the trajectory advances according to
\begin{equation}
    z_{\tau_{i+1}}
    =
    \mu_{\mathrm{ref},i}(z_{\tau_i},c)
    + \guidance_i\,
    \rms\!\left(\Delta_{\mathrm{ref},i}(z_{\tau_i},c)\right)\widehat g_i^{\mathcal M}.
    \label{eq:guided-update}
\end{equation}
Equation~\ref{eq:guided-update} is identical to the target construction in Eq.~\ref{eq:rg-opd-teacher}, but its role differs: RG-OPD distills the transition into $v_\theta$, whereas RGS uses it directly as the next latent. These equations show the Euler update used in our primary experiments. For a higher-order deterministic integrator, $\mu_{\mathrm{ref},i}$ is replaced by that integrator's output over $[\tau_i,\tau_{i+1}]$. The correction remains a single post-step update scaled by the resulting displacement.

\paragraph{Guidance scheduler.} Applying corrections at low noise levels brings marginal benefit and can distort fine details, so $\guidance_i$ follows a three-band schedule: $\guidance_{\mathrm{early}}$ for $\tau_i>\tau_{\mathrm{hi}}$, $\guidance_{\mathrm{mid}}$ for $\tau_{\mathrm{lo}}<\tau_i\le \tau_{\mathrm{hi}}$, and $0$ for $\tau_i\le \tau_{\mathrm{lo}}$, with $\guidance_{\mathrm{early}}>\guidance_{\mathrm{mid}}\ge0$. Algorithm~\ref{alg:reward-guided-sampling} outlines RGS. Backbone-specific thresholds and strengths are detailed in Appendix~\ref{sec:appendix-settings-guidance}.

\section{Experiments}
\label{sec:experiments}

We evaluate the framework along three axes: prediction accuracy of Latent Reward Registers on noisy latents, training efficiency of RG-OPD against RL and reward backpropagation baselines, and inference effectiveness of RGS relative to training-free sampling methods. Primary evaluations use SD3-Medium, with FLUX.1-dev providing cross-architecture evidence, and we then ablate register designs and analyze noise robustness. Full protocols and extended analyses appear in the Appendix.

\subsection{Latent Reward Prediction}
\label{sec:exp-accuracy}

\paragraph{Setup.}
We prepend $K{=}32$ register tokens to a frozen SD3-Medium backbone, training only the auxiliary branch. We measure pairwise preference accuracy across four benchmarks totaling $54{,}170$ pairs: ImageReward \citep{xu2023imagereward}, HPDv2 \citep{wu2023hpsv2}, HPDv3 \citep{ma2025hpsv3}, and GenAI-Bench \citep{li2024genaibench}. Accuracy is defined as the frequency with which a model assigns a higher score to the human-preferred image within a held-out pair. Latent reward models, including our method, evaluate paired latents forward-noised to the high-noise regime $t{=}0.8$. Standard endpoint models (ImageReward, PickScore, HPSv2, HPSv3) score the clean decoded images, providing a cross-regime reference. We compare against four families of latent reward baselines: LRM~\citep{zhang2025lrm} on SD1.5 and SDXL, PAVRM~\citep{mi2025videopavrm} (re-implemented on SD3-Medium), Diffusion Probe~\citep{cui2026diffusionprobegeneratedimage}, and DiNa-LRM~\citep{liu2026dinalrm}. For fair comparison, Table~\ref{tab:lrm-quality} matches our backbone to the SD3-Medium baselines; Appendix~\ref{sec:appendix-additional-register} details FLUX.1-dev results.

\paragraph{Results.}
The reward register predicts preferences accurately from noisy inputs: it outperforms all latent baselines on three of the four benchmarks (Table~\ref{tab:lrm-quality}) despite evaluating heavily corrupted latents. Clean-image reward models score uncorrupted inputs and are included as a reference point rather than a matched comparison. Appendix~\ref{sec:appendix-additional-register} shows that these prediction capabilities transfer to FLUX.1-dev.

\begin{table}[t]
\centering
\footnotesize
\setlength{\tabcolsep}{5pt}
\renewcommand{\arraystretch}{1.2}
\caption{\textbf{High-noise pairwise preference accuracy (\%) on four benchmarks.} Latent reward models score noisy states at $t{=}0.8$, whereas endpoint reward models score clean images. \best{Red bold} and \second{blue underline} mark the best and second-best results within each model category.}
\vspace{+0.2em}
\label{tab:lrm-quality}
\resizebox{\linewidth}{!}{%
\begin{tabular}{llccccc}
\toprule
 & & \multicolumn{5}{c}{\textbf{Pairwise Preference Accuracy (\%)}} \\ \cmidrule(lr){3-7}
\multirow{-2}{*}{\textbf{Model}} & \multirow{-2}{*}{\textbf{Backbone}} & \textbf{ImageReward} & \textbf{HPDv2} & \textbf{HPDv3} & \textbf{GenAI-Bench} & \textbf{Avg} \\ \midrule
\multicolumn{7}{@{}l}{\textbf{\textit{Endpoint Reward Models}}} \\
\quad ImageReward & CLIP & 65.15 & 73.95 & 58.74 & 63.41 & 65.31 \\
\quad PickScore  & CLIP & 62.73 & 79.44 & \second{65.67} & \second{69.98} & 69.45 \\
\quad HPSv2  & CLIP & \second{65.62} & \second{82.58} & 64.69 & 67.62 & \second{70.13} \\
\quad HPSv3 & Qwen2VL-7B & \best{67.03} & \best{85.36} & \best{76.03} & \best{70.95} & \best{74.84} \\
\midrule
\addlinespace[0.2em]

\multicolumn{7}{@{}l}{\textbf{\textit{Latent Reward Models}}} \\
\quad LRM-SD1.5 & SD-1.5 & 55.88 & 64.80 & 52.37 & 52.46 & 56.38 \\
\quad LRM-SDXL & SDXL    & 58.13 & 66.49 & 52.87 & 54.67 & 58.04 \\
\quad PAVRM & SD3-Medium & 56.76 & 73.52 & 68.18 & 57.20 & 63.91 \\
\quad Diffusion Probe & SD3-Medium & 58.67 & 75.29 & 68.68 & 57.14 & 64.95 \\
\quad DiNa-LRM & SD3-Medium & 60.10 & \best{78.76} & 68.08 & 58.94 & 66.47 \\
\rowcolor{highlightrow}
\textbf{Reward Register} & SD3-Medium & & & & & \\
\rowcolor{highlightrow}
\quad\texttt{+HPS} & & 61.84 & \second{78.46} & \best{69.96} & 61.41 & \second{67.92} \\
\rowcolor{highlightrow}
\quad\texttt{+ImageReward} & & \best{63.32} & 75.73 & 63.20 & \second{62.61} & 66.22 \\
\rowcolor{highlightrow}
\quad\texttt{+Raw score-sum} & & \second{63.12} & 78.10 & \second{69.10} & \best{62.95} & \best{68.32} \\
\bottomrule
\end{tabular}
}
\end{table}

\begin{figure}[!t]
    \centering
    \includegraphics[width=1\linewidth]{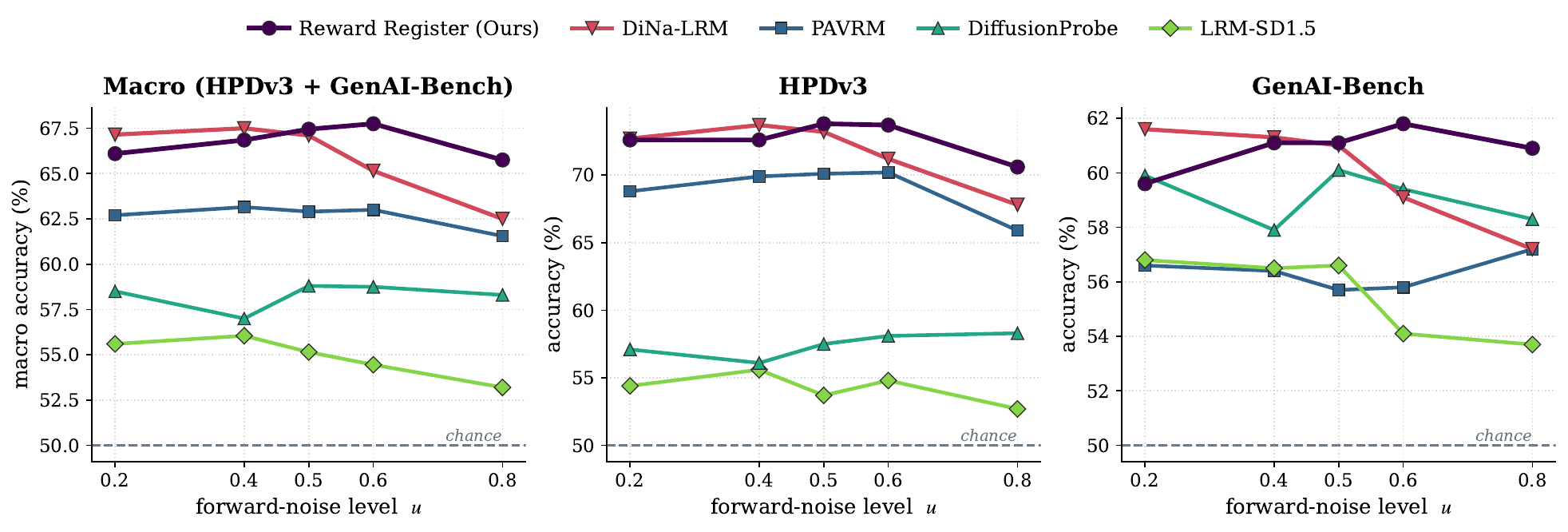}
    \caption{\textbf{Preference accuracy across noise levels.} Pairwise preference accuracy of latent reward models evaluated at noise level $t$. The reward register retains accuracy into the high-noise regime, whereas DiNa-LRM degrades sharply as the latents become less informative.}
    \label{fig:noise-robustness}
\end{figure}

\begin{table}[t]
\centering
\footnotesize
\setlength{\tabcolsep}{5pt}
\renewcommand{\arraystretch}{1.2}

\caption{\textbf{Training-time preference alignment on SD3-Medium and FLUX.1-dev.} HPSv3 and ImageReward are optimization targets, MUSIQ and CLIP-IQA measure no-reference perceptual quality. \best{Red bold} and \second{blue underline} mark the best and second-best results within each category.}
\vspace{+0.2em}
\label{tab:training-comparison}
\resizebox{\linewidth}{!}{%
\begin{tabular}{@{} l cccc cccc @{}}
\toprule
 & \multicolumn{4}{c}{\textbf{SD3-Medium}} & \multicolumn{4}{c}{\textbf{FLUX.1-dev}} \\
\cmidrule(lr){2-5}\cmidrule(l){6-9}
\textbf{Method} & HPSv3$\uparrow$ & IR$\uparrow$ & MUSIQ$\uparrow$ & CIQA$\uparrow$
                & HPSv3$\uparrow$ & IR$\uparrow$ & MUSIQ$\uparrow$ & CIQA$\uparrow$ \\
\midrule

\multicolumn{9}{@{}l}{\textbf{\textit{Baselines}}} \\
\quad Base + CFG & 8.55 & 1.183 & 73.2 & 0.694 & 9.89 & 1.142 & 72.9 & 0.669 \\
\quad ReFL \texttt{+HPS}      & 10.84 & 1.341 & 67.0 & 0.512 & 12.00 & 1.337 & 56.8 & 0.428 \\
\quad ReFL \texttt{+IR}       & 9.69 & \best{1.678} & \best{75.0} & \second{0.716} & 9.31 & 1.293 & 62.7 & 0.425 \\
\quad AlignProp \texttt{+HPS} & 11.20 & 1.472 & 68.1 & 0.580 & 11.83 & 1.241 & 59.4 & 0.481 \\
\quad AlignProp \texttt{+IR}  & 9.51 & \second{1.611} & 71.2 & 0.491 & 9.47 & 1.213 & 68.2 & 0.577 \\
\quad Flow-GRPO \texttt{+HPS} & 10.26 & 1.349 & 74.3 & 0.692 & \second{12.06} & 1.212 & \best{75.9} & 0.697 \\
\quad Flow-GRPO \texttt{+IR}  & 9.45 & 1.297 & 74.1 & 0.705 & 11.13 & \best{1.377} & \second{75.7} & \second{0.702} \\
\quad Diffusion-NFT \texttt{+HPS} & 9.51 & 0.995 & 72.4 & 0.551 & 11.54 & 1.056 & 75.1 & 0.695 \\
\quad Diffusion-NFT \texttt{+IR}  & 9.68 & 1.221 & 74.4 & 0.686 & 11.50 & 1.241 & 74.1 & 0.699 \\

\midrule
\addlinespace[0.2em]
\multicolumn{9}{@{}l}{\textbf{\textit{\method\ with on-policy distillation (Ours)}}} \\
\rowcolor{highlightrow}
\quad \method\ (OPD) \texttt{+HPS} & \second{11.57} & 1.336 & \second{74.9} & \best{0.726} & \best{12.38} & 1.229 & 74.4 & 0.648 \\
\rowcolor{highlightrow}
\quad \method\ (OPD) \texttt{+IR} & 10.98 & 1.444 & 72.1 & 0.674 & 10.82 & \second{1.367} & 75.6 & \best{0.704} \\
\rowcolor{highlightrow}
\quad \method\ (OPD) \texttt{+MH} & \best{11.62} & 1.417 & 74.0 & 0.704 & 11.91 & 1.346 & 75.0 & 0.681 \\
\bottomrule
\end{tabular}%
}
\end{table}

\begin{figure}[!t]
    \centering
    \includegraphics[width=\linewidth]{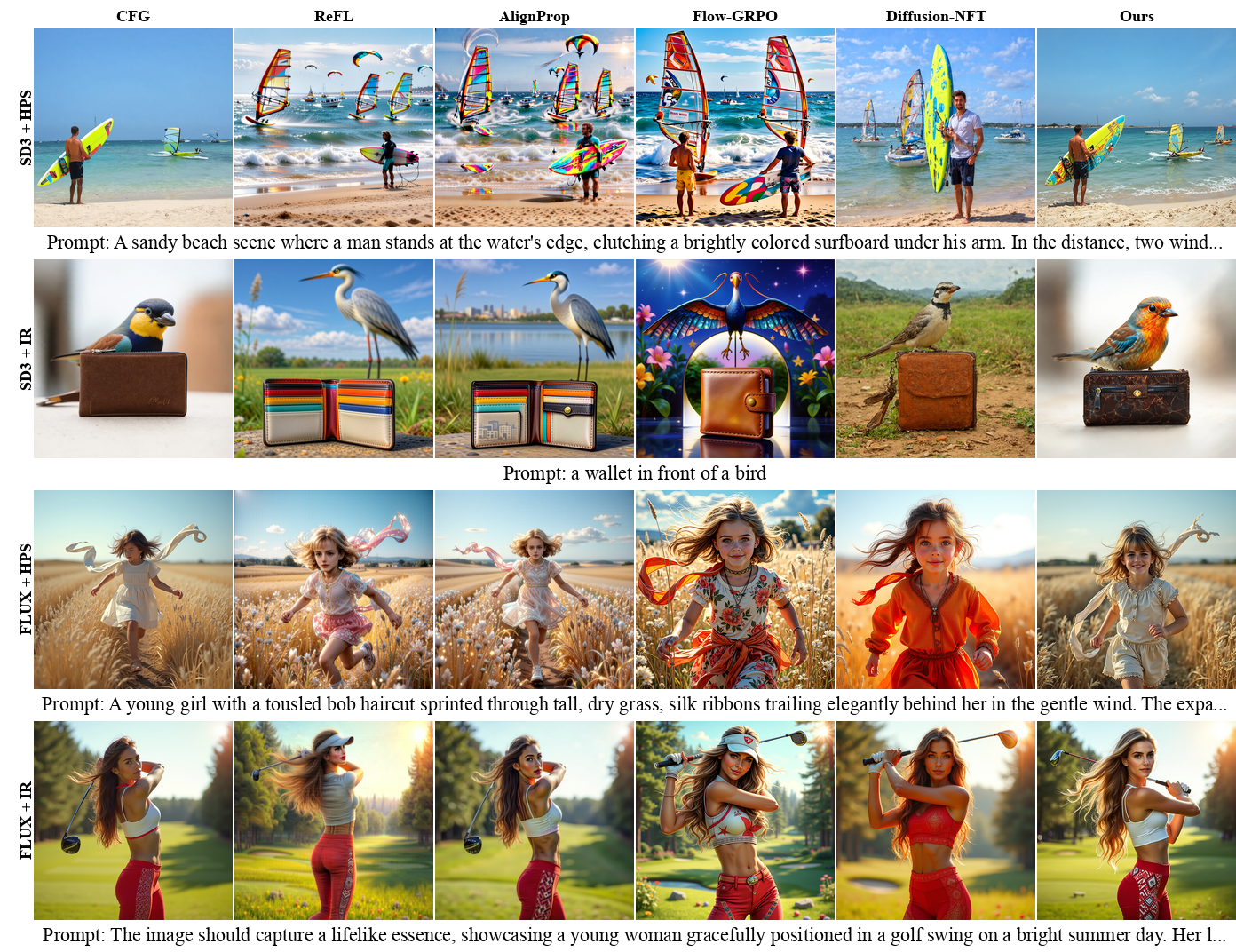}
    \vspace{-0.2cm}
    \caption{\textbf{Qualitative comparison of training-time alignment.} RG-OPD is compared against CFG, ReFL, AlignProp, Flow-GRPO, and Diffusion-NFT under matched prompts and seeds. RG-OPD achieves better prompt alignment and higher perceptual quality without introducing visible artifacts.}
    \vspace{-0.4cm}
    \label{fig:training-time-visual}
\end{figure}

\begin{table}[t]
\centering
\footnotesize
\setlength{\tabcolsep}{5pt}
\renewcommand{\arraystretch}{1.2}
\caption{\textbf{Reward-Guided Sampling with frozen SD3-Medium and FLUX.1-dev generators.} RGS provides the strongest HPS alignment and a favorable reward--quality trade-off among the evaluated training-free methods. HPSv3 and ImageReward (IR) are optimization targets. MUSIQ and CLIP-IQA (CIQA) measure perceptual quality. DNO and Demon use HPSv3 for their \texttt{+HPS} variants. \best{Red bold} and \second{blue underline} mark the best and second-best results within each backbone.}
\vspace{+0.2em}
\label{tab:guided-comparison}
\resizebox{\linewidth}{!}{%
\begin{tabular}{@{} l cccc cccc @{}}
\toprule
 & \multicolumn{4}{c}{\textbf{SD3-Medium}} & \multicolumn{4}{c}{\textbf{FLUX.1-dev}} \\
\cmidrule(lr){2-5}\cmidrule(l){6-9}
\textbf{Method} & HPSv3$\uparrow$ & IR$\uparrow$ & MUSIQ$\uparrow$ & CIQA$\uparrow$
                & HPSv3$\uparrow$ & IR$\uparrow$ & MUSIQ$\uparrow$ & CIQA$\uparrow$ \\
\midrule
\multicolumn{9}{@{}l}{\textbf{\textit{Baselines}}} \\
\quad Base + CFG & 8.55 & 1.183 & 73.2 & 0.694 & 9.89 & 1.142 & 72.9 & 0.669 \\
\quad DNO \texttt{+HPS}   & 8.85 & 1.272 & 73.1 & 0.698 & 9.64 & 1.191 & 71.9 & 0.655 \\
\quad Demon \texttt{+HPS} & 9.09 & 1.311 & 73.0 & 0.685 & 9.56 & 1.209 & 68.2 & 0.605 \\
\quad DNO \texttt{+IR}    & 8.19 & 1.454 & 72.3 & 0.693 & 9.25 & \second{1.350} & 69.5 & 0.632 \\
\quad Demon \texttt{+IR}  & 8.90 & \best{1.596} & 72.5 & 0.674 & 9.27 & \best{1.501} & 68.2 & 0.603 \\
\midrule
\addlinespace[0.2em]
\multicolumn{9}{@{}l}{\textbf{\textit{\samplingmethod\ (Ours)}}} \\
\rowcolor{highlightrow}
\quad \method\ \texttt{+HPS} & \best{10.70} & 1.347 & \best{75.0} & \best{0.728} & \best{11.03} & 1.212 & \second{73.3} & \second{0.715} \\
\rowcolor{highlightrow}
\quad \method\ \texttt{+IR}  & 10.12 & \second{1.465} & 73.0 & 0.694 & 10.38 & 1.297 & 73.2 & 0.710 \\
\rowcolor{highlightrow}
\quad \method\ \texttt{+MH}  & \second{10.48} & 1.413 & \second{74.1} & \second{0.723} & \second{10.80} & 1.276 & \best{73.5} & \best{0.717} \\
\bottomrule
\end{tabular}%
}
\end{table}

\begin{figure*}[t]
\centering
\includegraphics[width=\linewidth]{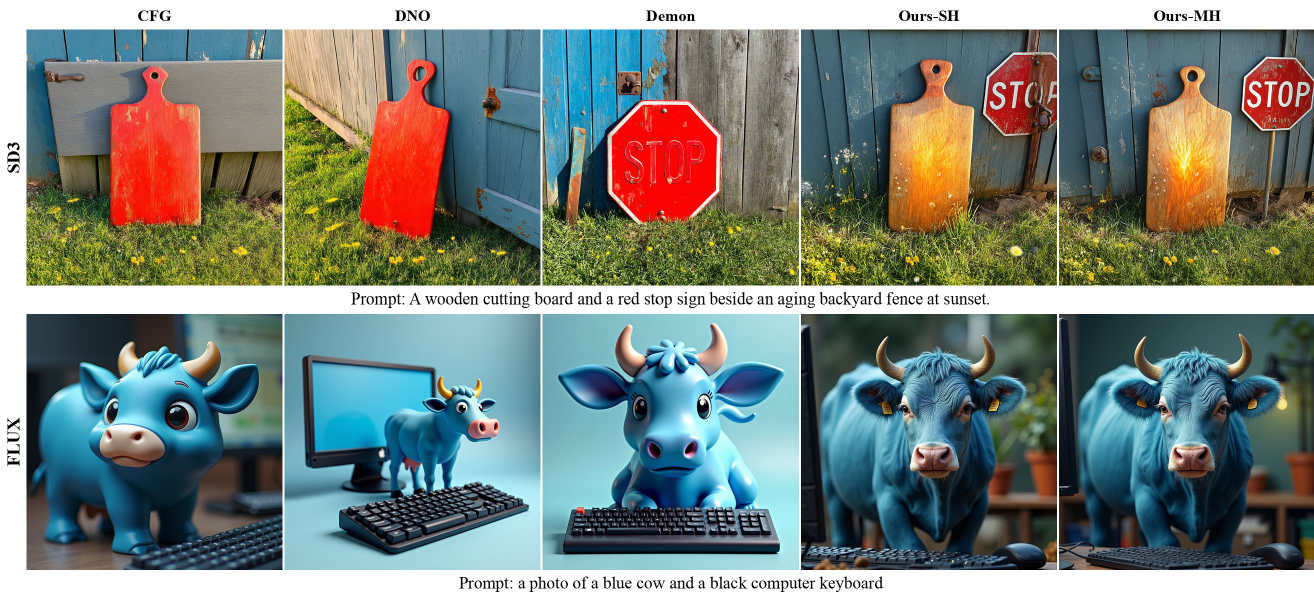}
\vspace{-0.4cm}
\caption{\textbf{Comparison of training-free sampling methods.} CFG, DNO, Demon, and our single- and multi-head RGS variants use matched prompts and seeds within each backbone.}
\vspace{-0.4cm}
\label{fig:test-time-guided-comparison}
\end{figure*}

\subsection{Reward-Gradient On-Policy Distillation}
\label{sec:exp-training}

\paragraph{Setup.}
We benchmark RG-OPD against reward-backpropagation (ReFL, AlignProp) and online RL methods (Flow-GRPO, Diffusion-NFT). All approaches update the generator weights. Evaluations utilize matched prompts, seeds, resolutions, and metric pipelines. HPSv3 and ImageReward serve as the optimization targets, while MUSIQ and CLIP-IQA assess reward-independent perceptual quality. Appendix~\ref{sec:appendix-settings-opd} details the complete training protocol.

\paragraph{Results.}
RG-OPD attains the highest HPSv3 scores on both architectures while keeping a balanced multi-reward profile (Table~\ref{tab:training-comparison}). Reward-backpropagation baselines, by contrast, often maximize target metrics at the expense of visible perceptual degradation, a classic signature of reward over-optimization. RG-OPD instead realizes a favorable reward--quality trade-off. Figure~\ref{fig:training-time-visual} shows this qualitatively: under matched prompts and seeds, RG-OPD preserves prompt-relevant content and perceptual quality without the artifacts exhibited by several baselines.

\subsection{Reward-Guided Sampling}
\label{sec:exp-guidance}

\paragraph{Setup.}
We compare RGS with the training-free baselines DNO and Demon under matched prompts, seeds, resolutions, and metrics. All methods keep the generator frozen; DNO and Demon optimize or search per sample using external reward models, whereas RGS uses analytical gradients from the latent reward register at each active step. HPSv3 and ImageReward quantify target alignment, and MUSIQ \citep{ke2021musiq} and CLIP-IQA \citep{wang2023exploring} measure no-reference perceptual quality. Extended protocol details are given in Appendix~\ref{sec:appendix-settings-guidance}.

\paragraph{Results.}
Each single-head RGS variant improves its target metric relative to the CFG baseline while preserving perceptual quality, and RGS reaches the highest HPSv3 score among all evaluated training-free methods, with a favorable reward--quality trade-off under ImageReward optimization. Figure~\ref{fig:test-time-guided-comparison} gives the corresponding qualitative comparison; prompts and seeds are matched to hold input conditions constant, so differences in prompt alignment and visual quality are not explained by prompt or seed selection.

Memory is another advantage of RGS. Because it shares the frozen generator backbone, peak memory stays comparable to unguided CFG and well below that of guided baselines, which require separate external scorer networks (Table~\ref{tab:efficiency}).

\subsection{Analysis}
\label{sec:exp-analysis}

\paragraph{Register design.}
The persistent register state is the main driver of accuracy. Removing the register causes the largest performance drop, while the skip-FFN design and the trainable query projection each add complementary gains (Table~\ref{tab:ablation-design}). Reducing the number of readout layers, by contrast, has little effect, indicating that persistent side-stream aggregation matters more than deep multi-scale feature extraction.

\begin{table}[t]
\centering
\small
\renewcommand{\arraystretch}{1.15}
\caption{\textbf{Reward Register Design Ablation.} Pairwise preference accuracy evaluated across four benchmarks. $\Delta$ quantifies the deviation in average accuracy relative to the full deployed method. \textbf{Bold} indicates the optimal performance within each respective column.}
\label{tab:ablation-design}
\begin{tabular*}{\linewidth}{@{\extracolsep{\fill}} l ccccc c @{}}
\toprule
\multirow{2}{*}{\textbf{Variant}} & \multicolumn{5}{c}{\textbf{Pairwise Preference Accuracy (\%)}} & \multirow{2}{*}{$\Delta$} \\
\cmidrule(lr){2-6}
 & \textbf{ImageReward} & \textbf{HPDv2} & \textbf{HPDv3} & \textbf{GenAI-Bench} & \textbf{Avg} & \\
\midrule
\textbf{Full method} & \textbf{63.12} & 78.10 & 69.10 & \textbf{62.95} & \textbf{68.32} & \textbf{---} \\
\midrule
\quad w/o reward register   & 56.86 & 74.12 & 65.42 & 60.76 & 64.29 & $-4.03$ \\
\quad w/o skip-FFN          & 58.15 & 77.82 & 68.54 & 60.38 & 66.23 & $-2.09$ \\
\quad w/o trainable q-proj  & 58.20 & 77.62 & 68.62 & 60.42 & 66.21 & $-2.11$ \\
\midrule
\quad w/ layers $[4]$       & 59.39 & 79.32 & 69.57 & 61.69 & 67.49 & $-0.83$ \\
\quad w/ layers $[4,8]$      & 60.01 & \textbf{79.90} & \textbf{70.35} & 61.30 & 67.89 & $-0.43$ \\
\bottomrule
\end{tabular*}
\end{table}

\begin{figure*}[t]
    \centering
    \includegraphics[width=\textwidth]{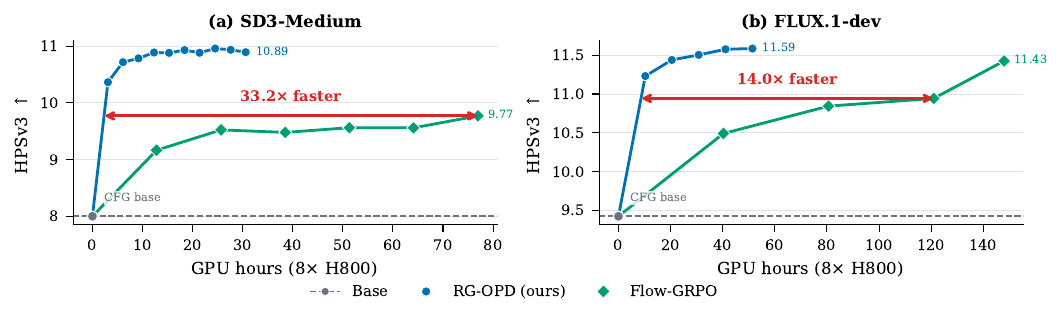}
    \vspace{-0.4cm}
    \caption{\textbf{Training efficiency of RG-OPD.} HPSv3 optimization trajectories of RG-OPD and Flow-GRPO against aggregate GPU hours on SD3-Medium and FLUX.1-dev. RG-OPD reaches equivalent HPSv3 score while $14.0\times$--$33.2\times$ faster than Flow-GRPO in GPU hours, reflecting the dense, step-wise supervision provided by the latent reward registers.}
    \label{fig:training-efficiency}
    \vspace{-0.3cm}
\end{figure*}

\begin{figure*}[t]
    \centering
    \includegraphics[width=\textwidth]{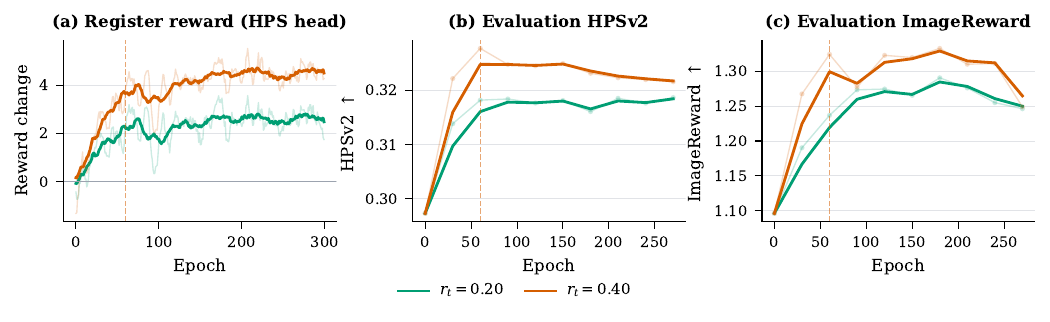}
    \vspace{-0.7cm}
    \caption{\textbf{Effect of the reward tilt $\guidance$ during RG-OPD training.} Training dynamics are compared between $\guidance{=}0.20$ and $\guidance{=}0.40$ (denoted $r_t$ in the panel labels). The results show that a larger reward scale accelerates convergence and yields a higher peak performance. Moreover, the upward trend of the internal latent reward (a) remains consistent with the endpoint HPSv2 evaluation (b), indicating alignment between the intermediate proxy and the terminal objective. ImageReward (c) is evaluated as an independent, unoptimized metric to monitor the overall generation quality during the process.}
    \label{fig:teacher-vs-eval}
\end{figure*}

\paragraph{Training efficiency.}
Dense reward gradients provide continuous supervision throughout the denoising process, which accelerates convergence substantially relative to policy-gradient RL. On-policy trajectories meanwhile track the evolving student distribution. As Figure~\ref{fig:training-efficiency} shows, RG-OPD reaches Flow-GRPO-equivalent HPSv3 levels $14.0\times$--$33.2\times$ faster in GPU hours on both backbones. The efficiency curves are evaluated on 256 prompts to track training progress, while the main results in Table~\ref{tab:training-comparison} use the full 400/800 prompt--seed pairs, so the two evaluations report somewhat different absolute scores. Figure~\ref{fig:teacher-vs-eval} further compares training dynamics under two reward tilts ($\guidance{=}0.20$ vs.\ $0.40$): the larger tilt accelerates convergence and raises peak performance, and the internal register reward tracks the endpoint HPSv2 evaluation throughout training.

\begin{figure}[t]
    \centering
    \includegraphics[width=1\linewidth]{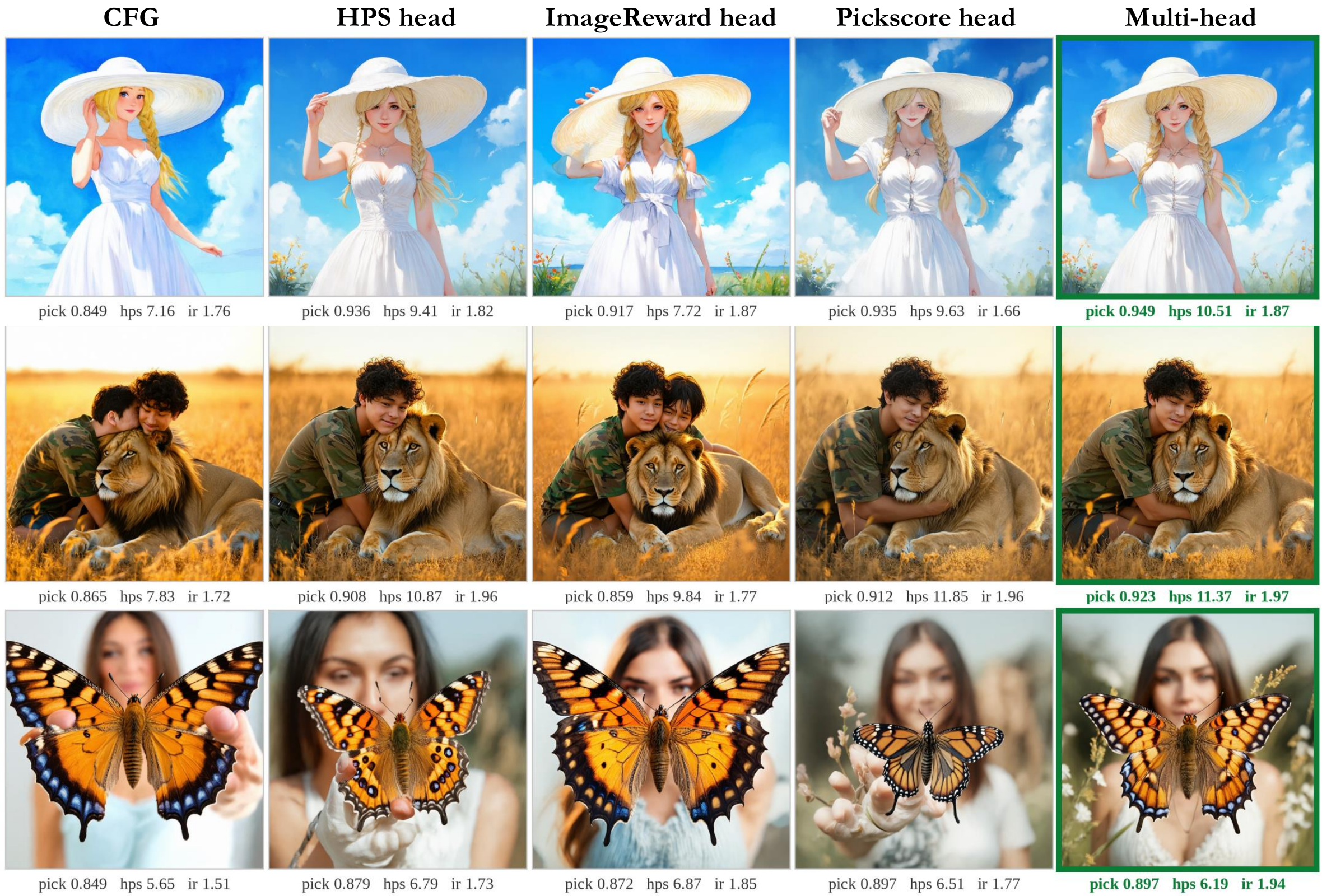}
    \vspace{-0.4cm}
    \caption{\textbf{Multi-head RGS.} Multi-head reward gradients achieve a balanced performance across different reward metrics, compared to single-head RGS under identical settings.}
    \vspace{-0.5cm}
    \label{fig:head-compare}
\end{figure}

\paragraph{Multi-head RGS.}
Multi-reward RGS balances competing objectives. Where single-head guidance targets one reward, the multi-head variant maintains a balanced profile across reward and perceptual metrics, consistent with the trends in Table~\ref{tab:guided-comparison} and Figure~\ref{fig:head-compare}. This supports the effectiveness of the equal-RMS aggregation strategy for multi-objective composition.

\paragraph{Accuracy across noise levels.}
\textbf{The register stays reliable at high noise levels.} DiNa-LRM performs competitively at low noise but degrades sharply as the latent inputs become uninformative, whereas the register retains accuracy throughout (Figure~\ref{fig:noise-robustness}). This robustness matters precisely in the high-noise regime used for early guidance, where the register acts as a continuous step-wise signal rather than a late-stage endpoint proxy.

\paragraph{Additional analyses.}
Appendix~\ref{sec:appendix-additional-test} demonstrates that alignment improvements persist across varying sampling budgets, guidance strengths, and deterministic ODE solvers. It also analyzes the temporal accumulation of reward corrections along the generated trajectory. Appendix~\ref{sec:appendix-additional-register} empirically validates that the constituent register heads preserve the relative ranking distributions established by their corresponding endpoint reward models.

\section{Conclusion}
\label{sec:conclusion}
In this work, we demonstrate that intermediate DiT representations capture terminal reward information early in the denoising process. Building on this insight, we propose Latent Reward Registers, a non-intrusive auxiliary stream that predicts endpoint rewards directly from intermediate noisy latents. The resulting dense reward field supports two alignment strategies. Reward-Gradient On-Policy Distillation (RG-OPD) utilizes step-wise gradient supervision to outperform online RL baselines in efficiency and alignment, while avoiding the over-optimization typical of direct reward backpropagation. Furthermore, Reward-Guided Sampling (RGS) applies magnitude-matched register gradients to steer generation without parameter updates, improving multi-objective target rewards while preserving perceptual quality. Ultimately, by resolving the credit assignment problem, this framework establishes a reliable latent reward model for diffusion RL.

\bibliographystyle{iclr2026_conference}
\bibliography{rt_guided_sampling_iclr2027}

\appendix
\newpage

\section{Appendix Overview}
\label{sec:appendix-overview}

This appendix contains supplementary material: the experimental protocols, extended results for Latent Reward Registers, training-time alignment, and Reward-Guided Sampling (RGS), and directions for future research.

\begin{enumerate}[nosep, leftmargin=2em]
    \item \textbf{Experimental Settings} (Appendix~\ref{sec:appendix-settings})
    \begin{itemize}[nosep, leftmargin=1.5em]
        \item Reward register training (Appendix~\ref{sec:appendix-settings-register})
        \item Reward-gradient on-policy distillation (Appendix~\ref{sec:appendix-settings-opd})
        \item Reward-Guided Sampling (Appendix~\ref{sec:appendix-settings-guidance})
    \end{itemize}
    \item \textbf{Additional Reward-Register Results} (Appendix~\ref{sec:appendix-additional-register})
    \begin{itemize}[nosep, leftmargin=1.5em]
        \item Pairwise preference accuracy (Appendix~\ref{sec:appendix-register-accuracy})
        \item Rank agreement with endpoint reward models (Appendix~\ref{sec:appendix-rank-agreement})
    \end{itemize}
    \item \textbf{Additional Training-Time Alignment Results} (Appendix~\ref{sec:appendix-additional-train})
    \begin{itemize}[nosep, leftmargin=1.5em]
        \item Qualitative results (Appendix~\ref{sec:appendix-train-qualitative})
    \end{itemize}
    \item \textbf{Additional Reward-Guided Sampling Results} (Appendix~\ref{sec:appendix-additional-test})
    \begin{itemize}[nosep, leftmargin=1.5em]
        \item Magnitude-matched CFG-direction control (Appendix~\ref{sec:appendix-cfg-direction-control})
        \item Inference cost (Appendix~\ref{sec:appendix-cost})
        \item Sampling budgets and guidance strengths (Appendix~\ref{sec:appendix-step-scale})
        \item Reward along the guided trajectory (Appendix~\ref{sec:appendix-reward-trajectory})
        \item Robustness to the ODE solver (Appendix~\ref{sec:appendix-ode-solver})
        \item Qualitative comparison across prompts (Appendix~\ref{sec:gallery})
    \end{itemize}
    \item \textbf{Future Work} (Appendix~\ref{sec:appendix-future})
\end{enumerate}

\section{Experimental Settings}
\label{sec:appendix-settings}

\paragraph{Reward Register Training Setting}
\label{sec:appendix-settings-training}
\label{sec:appendix-settings-register}
Table~\ref{tab:register-training-config} lists the hyperparameters and configuration details for training the reward register that were omitted from the main text, distinguishing backbone-specific architectural choices from the shared optimization settings.

\paragraph{Reward-Gradient On-Policy Distillation Setting}
\label{sec:appendix-settings-opd}
Table~\ref{tab:opd-training-config} lists the optimization configurations and hyperparameters for SD3-Medium and FLUX.1-dev. Direct epoch-to-epoch comparisons between RG-OPD and the online RL baselines would be misleading, since the two paradigms consume data at different rates. The main text therefore reports computational efficiency in aggregate GPU hours.

\begin{table}[b]
\centering
\footnotesize
\renewcommand{\arraystretch}{1.12}
\setlength{\tabcolsep}{4pt}
\caption{\textbf{RG-OPD optimization details.} Model checkpoints were selected using a validation set disjoint from the final test set. Computational budgets are reported in terms of GPU hours and optimizer updates to facilitate fair comparison across paradigms with varying rollout densities.}
\label{tab:opd-training-config}
\begin{tabularx}{\linewidth}{@{}CCC@{}}
\toprule
\textbf{Setting} & \textbf{SD3-Medium Configuration} & \textbf{FLUX.1-dev Configuration} \\
\midrule
Student adaptation & LoRA; merged for evaluation & LoRA; merged for evaluation \\
Training length & 60--100 epochs & 150 epochs \\
Training rollout & 10 FlowMatch Euler steps & 10 FlowMatch Euler steps \\
Optimized denoising steps $\mathcal{J}$ & First 9 of 10 rollout steps ($\rho=0.99$) & First 9 of 10 rollout steps ($\rho=0.99$) \\
Teacher reward tilt $\guidance_i$ & $\{0.20, 0.40, 0.50\}$ for $\tau_i{>}0.2$; 0 otherwise & $0.80$ for $\tau_i{>}0.2$; 0 otherwise \\
\midrule
Flow-GRPO / Diffusion-NFT & LoRA $r{=}32/\alpha{=}64$; 6-step SDE; 100 epochs & LoRA $r{=}32/\alpha{=}64$; 6-step SDE; 100 epochs \\
Rollout samples (RL baselines) & $1{,}152$ rollout images per epoch & $1{,}152$ rollout images per epoch \\
\bottomrule
\end{tabularx}
\end{table}

\paragraph{Reward-Guided Sampling Setting}
\label{sec:appendix-settings-guidance}
Table~\ref{tab:sampling-config} documents the sampling configurations and guidance hyperparameters omitted from the main text.

\section{Additional Reward-Register Results}
\label{sec:appendix-additional-register}
\label{sec:appendix-settings-accuracy}

\subsection{Pairwise Preference Accuracy}
\label{sec:appendix-register-accuracy}
Pairwise preference accuracy measures how often a model scores the human-preferred image higher within a held-out pair. The evaluation spans the ImageReward, HPDv2, HPDv3, and GenAI-Bench datasets, with $54{,}170$ pairs in total. Endpoint reward models are evaluated on fully decoded, pristine images, while latent reward models score pairs that have been forward-noised to $t{=}0.8$. Across both backbones, \texttt{+Raw score-sum} denotes an unweighted ensemble of the constituent register heads.

\begin{table}[t]
\centering
\footnotesize
\renewcommand{\arraystretch}{1.12}
\setlength{\tabcolsep}{3.5pt}
\caption{\textbf{Reward-register training details.} Backbone-specific architectural choices and shared optimization hyperparameters. Values spanning both columns are shared across architectures.}
\label{tab:register-training-config}
\begin{tabularx}{\linewidth}{@{}lCC@{}}
\toprule
\textbf{Setting} & \textbf{SD3-Medium} & \textbf{FLUX.1-dev} \\
\midrule
Register-equipped blocks $L$ & 12 (of 38) & 8 (of 19 double blocks) \\
Feature-snapshot blocks $\mathcal{B}$ & $\{4,8,12\}$ & $\{2,5,8\}$ \\
\midrule
Register tokens $K$ & \multicolumn{2}{c}{32} \\
Image-token pooling & \multicolumn{2}{c}{$64\times64$ to $16\times16$} \\
Readout attention heads & \multicolumn{2}{c}{8} \\
Endpoint reward objectives & \multicolumn{2}{c}{HPSv3, ImageReward} \\
Training data & \multicolumn{2}{c}{$30{,}652$ prompt groups, 4 images/group} \\
Validation data & \multicolumn{2}{c}{300 prompt groups} \\
Data split & \multicolumn{2}{c}{$98/1/1$ by prompt} \\
Training length & \multicolumn{2}{c}{3 epochs} \\
Batch size & \multicolumn{2}{c}{64} \\
Precision / optimizer & \multicolumn{2}{c}{bf16 AdamW} \\
Learning rate & \multicolumn{2}{c}{$5\times10^{-5}$} \\
Weight decay & \multicolumn{2}{c}{$0.01$} \\
Warmup / schedule & \multicolumn{2}{c}{1,000 steps / constant} \\
Gradient clipping & \multicolumn{2}{c}{$1.0$} \\
EMA decay & \multicolumn{2}{c}{$0.999$} \\
\bottomrule
\end{tabularx}
\end{table}

\begin{table}[t]
\centering
\footnotesize
\renewcommand{\arraystretch}{1.12}
\setlength{\tabcolsep}{3.5pt}
\caption{\textbf{Reward-Guided Sampling details.} Both SD3-Medium and FLUX.1-dev employ identical flow-time bands, with backbone-specific sampling step counts and mid-band guidance magnitudes.}
\vspace{+0.2em}
\label{tab:sampling-config}
\begin{tabularx}{\linewidth}{@{}lCC@{}}
\toprule
\textbf{Setting} & \textbf{SD3-Medium} & \textbf{FLUX.1-dev} \\
\midrule
Resolution & $1024 \times 1024$ & $1024 \times 1024$ \\
Solver & FlowMatch Euler & FlowMatch Euler \\
Sampling steps & 42 & 40 \\
Text guidance & CFG scale 4.5 & Embedded guidance 3.5 \\
\midrule
Early band ($\tau_i>0.8$) & $\guidance_{\mathrm{early}}=0.30$ & $\guidance_{\mathrm{early}}=0.30$ \\
Mid band ($0.2<\tau_i\leq0.8$) & $\guidance_{\mathrm{mid}}=0.05$ & $\guidance_{\mathrm{mid}}=0.10$ \\
Low-noise tail ($\tau_i\leq0.2$) & 0 & 0 \\
Multi-head combination & \multicolumn{2}{c}{Equal-weight sum of unit-RMS per-head gradients} \\
Correction mode & \multicolumn{2}{c}{Additive; no spatial gradient filtering} \\
\bottomrule
\end{tabularx}
\end{table}

\begin{table}[t]
\centering
\small
\setlength{\tabcolsep}{5pt}
\renewcommand{\arraystretch}{1.15}
\caption{\textbf{High-noise pairwise preference accuracy (\%) on FLUX.1-dev.} Register variants on a frozen FLUX.1-dev backbone, evaluated at $t{=}0.8$ under the protocol of Table~\ref{tab:lrm-quality}. \textbf{Bold} marks the best result in each column.}
\vspace{+0.2em}
\label{tab:lrm-quality-flux}
\begin{tabular}{lccccc}
\toprule
\textbf{Variant} & \textbf{ImageReward} & \textbf{HPDv2} & \textbf{HPDv3} & \textbf{GenAI-Bench} & \textbf{Avg} \\
\midrule
\texttt{+HPS} & 62.12 & 81.18 & \textbf{73.30} & 61.75 & 69.59 \\
\texttt{+ImageReward} & 63.13 & 78.84 & 65.86 & \textbf{63.42} & 67.81 \\
\texttt{+Raw score-sum} & \textbf{63.46} & \textbf{81.20} & 72.36 & 62.96 & \textbf{70.00} \\
\bottomrule
\end{tabular}
\end{table}

On FLUX.1-dev the raw score-sum ensemble reaches an average accuracy of $70.00\%$ (Table~\ref{tab:lrm-quality-flux}), confirming that the register mechanism transfers to an alternative frozen backbone.

\subsection{Rank Agreement with Endpoint Reward Models}
\label{sec:appendix-rank-agreement}

\begin{figure}[t]
    \centering
    \includegraphics[width=1\linewidth]{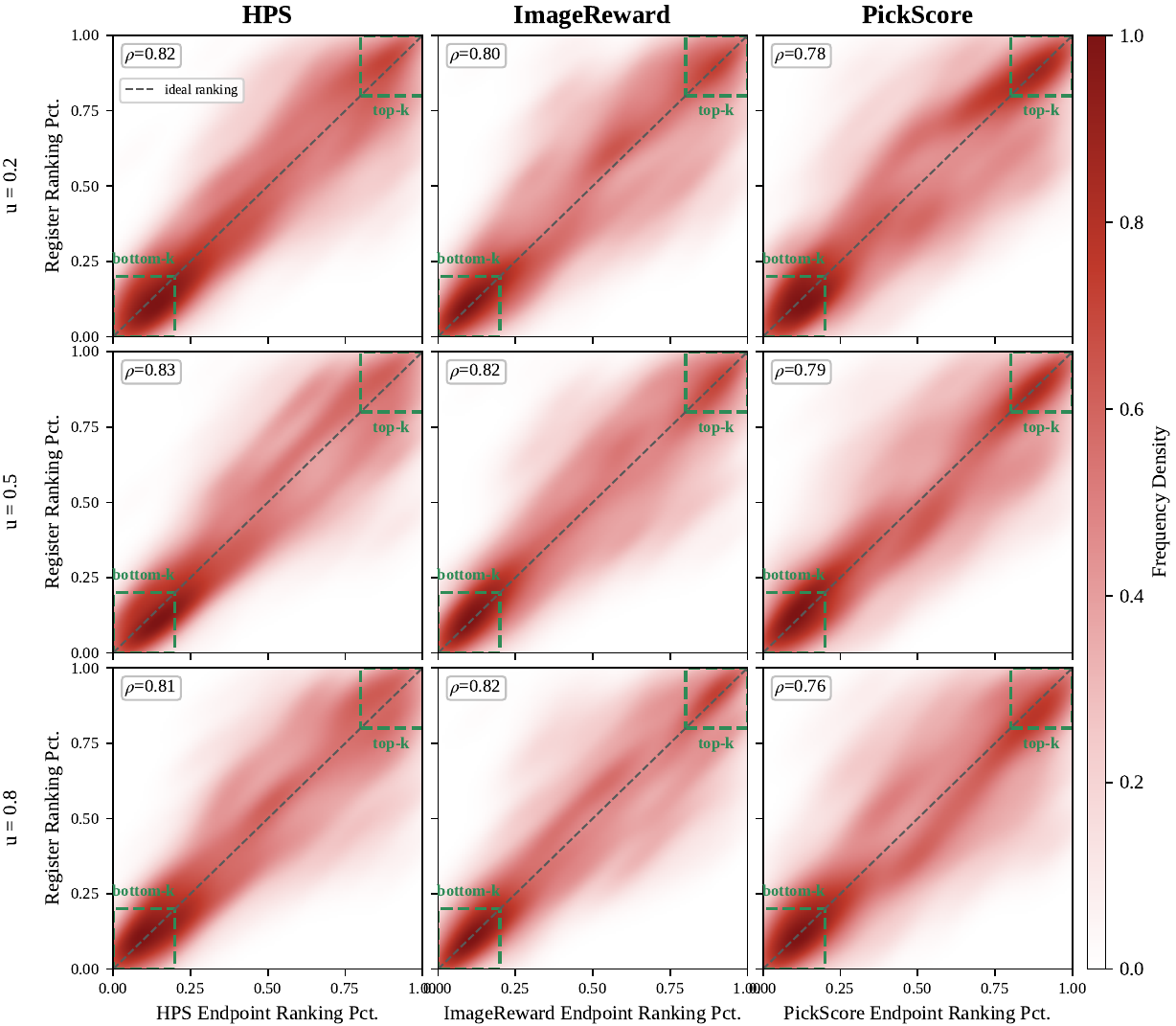}
    \caption{\textbf{Rank agreement between register heads and endpoint reward models.} Joint density of ranking percentiles from register heads on noised latents versus endpoint models on fully decoded images, across noise levels $t\in\{0.2,0.5,0.8\}$. $\rho$ denotes the Spearman rank correlation. The diagonal marks perfect rank preservation.}
    \label{fig:rank-agreement}
\end{figure}

\textbf{Each register head preserves the ranking behavior of its corresponding endpoint reward model at high noise.} Figure~\ref{fig:rank-agreement} compares the ranking percentiles the register assigns to noised latents with those of the endpoint models on fully denoised images, over a sample of $800$ images. The Spearman correlation $\rho$ ranges from $0.76$ to $0.83$ and stays stable across noise levels $t \in \{0.2, 0.5, 0.8\}$.

\section{Additional Training-Time Alignment Results}
\label{sec:appendix-additional-train}

\subsection{Qualitative Results}
\label{sec:appendix-train-qualitative}

\begin{figure}[p]
    \centering
    \includegraphics[width=0.90\linewidth]{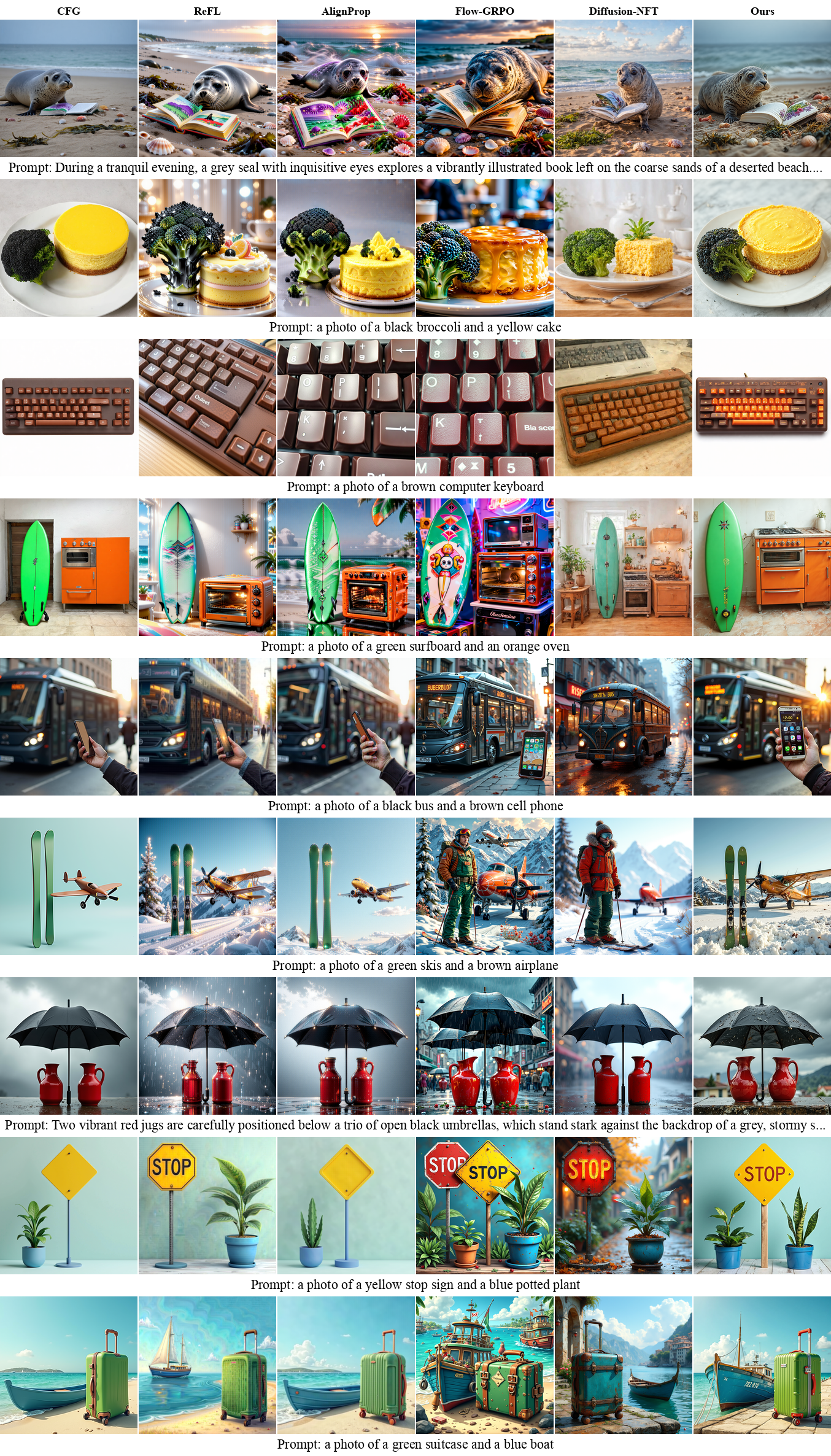}
    \caption{\textbf{Extended qualitative comparison of training-time alignment.} We provide additional visual examples comparing our RG-OPD framework against baseline methods across a diverse set of text prompts.}
    \label{fig:appendix-train-time}
\end{figure}

Figure~\ref{fig:appendix-train-time} provides additional qualitative examples of training-time alignment. Each row uses the same text prompt and initial latent seed, so the comparison reflects consistent gains in prompt adherence and visual quality across prompts.

\section{Additional Reward-Guided Sampling Results}
\label{sec:appendix-additional-test}

\subsection{Magnitude-Matched CFG-Direction Control}
\label{sec:appendix-cfg-direction-control}

To separate the effect of correction magnitude from correction direction, the CFG Control baseline replaces the RGS reward gradient with the SD3 conditional-unconditional residual. At each active step the applied correction is $\guidance_i\rms(\Delta_{\mathrm{ref},i})\unitrms(\Delta\tau_i\,(v_{\mathrm{cond}}-v_{\mathrm{uncond}}))$, following the same $\guidance_i$ schedule as RGS. The two controls therefore share the per-step magnitude-matching rule and differ only in the direction applied.

CFG Control improves some perceptual and reward metrics relative to standard CFG but does not approach the alignment of RGS (Table~\ref{tab:cfg-direction-control}). Since the correction magnitude is identical, this comparison isolates the learned reward direction as the driver of the improvements.

\begin{table}[h]
\centering
\small
\renewcommand{\arraystretch}{1.15}
\setlength{\tabcolsep}{7pt}
\caption{\textbf{Magnitude-matched direction control on SD3-Medium.} CFG Control matches the per-step RMS magnitude of the RGS correction but orients the update along the standard CFG residual instead of the learned reward direction. Evaluation follows the matched prompt--seed protocol of Table~\ref{tab:guided-comparison}. \best{Red bold} and \second{blue underline} mark the best and second-best results in each column.}
\vspace{+0.2em}
\label{tab:cfg-direction-control}
\begin{tabular}{l cccc}
\toprule
\textbf{Method} & \textbf{HPSv3}$\uparrow$ & \textbf{IR}$\uparrow$ & \textbf{MUSIQ}$\uparrow$ & \textbf{CIQA}$\uparrow$ \\
\midrule
Base + CFG & 8.55 & 1.183 & 73.2 & 0.694 \\
CFG Control & 8.50 & 1.214 & \second{74.14} & 0.720 \\
\midrule
\method\ \texttt{+HPS} & \best{10.70} & 1.347 & \best{75.0} & \best{0.728} \\
\method\ \texttt{+IR} & 10.12 & \best{1.465} & 73.00 & 0.694 \\
\method\ \texttt{+MH} & \second{10.48} & \second{1.413} & 74.10 & \second{0.723} \\
\bottomrule
\end{tabular}
\end{table}

\subsection{Inference Cost of Reward-Guided Sampling}
\label{sec:appendix-cost}

\begin{table}[t]
\centering
\small
\renewcommand{\arraystretch}{1.15}
\setlength{\tabcolsep}{2pt}
\caption{\textbf{Inference cost of RGS on SD3-Medium.} Comparison of trainable parameter budgets, memory footprints, and computational overhead against existing guided baselines. SD3 with standard CFG is provided as the unguided reference. $\Delta$FLOP/step denotes the marginal computational overhead relative to a base CFG step. The \textit{early-only} configuration applies guidance exclusively during the first 18 of 42 steps (where $\tau_i>0.8$). \textbf{Bold} marks the best efficiency among guided methods (lower is better).}
\vspace{+0.2em}
\label{tab:efficiency}
\resizebox{\linewidth}{!}{%
\begin{tabular}{l ccccc}
\toprule
\textbf{Method} & \textbf{Params}$\downarrow$ & \textbf{Peak mem}$\downarrow$ & \textbf{Reward Eval./step}$\downarrow$ & \textbf{$\Delta$FLOP/step}$\downarrow$ & \textbf{Latency/img}$\downarrow$ \\
 & (M) & (GiB) & (ms) & (TFLOP) & (s, $\times$CFG) \\
\midrule
\multicolumn{6}{l}{\textit{Reference (no reward)}} \\
SD3 + CFG       & --- & 16.9 & --- & 12.4 (base) & 2.99 \,(1.00$\times$) \\
\midrule
\multicolumn{6}{l}{\textit{Guided baselines (separate reward backbone)}} \\
PAVRM              & 691.4 & 31.3 & \textbf{40}  & $\mathbf{+4.1}$  & 4.98 \,(1.67$\times$) \\
Diffusion Probe    & \textbf{15.3}  & 34.3 & 108 & $+11.5$ & 7.85 \,(2.63$\times$) \\
Decode+ImageReward & 446.7 & 27.3 & 215 & $+10.2$ & 12.32 \,(4.12$\times$) \\
\midrule
\multicolumn{6}{l}{\textit{\method\ (frozen generator, reward register)}} \\
\method\ \texttt{+HPS} & 147.6 & \textbf{17.6} & 50 & $+5.9$ & 5.38 \,(1.80$\times$) \\
\method\ \texttt{+HPS} \textit{(early-only)} & 147.6 & \textbf{17.6} & 50 & $+5.9$ & \textbf{4.00} \,(\textbf{1.34}$\times$) \\
\bottomrule
\end{tabular}%
}
\end{table}

\textbf{\method\ has a memory footprint comparable to standard CFG and an inference latency competitive with the most efficient guided baselines (Table~\ref{tab:efficiency}).} All reported numbers are measured end-to-end on a single uncontended H800 GPU in bf16, generating $1024^2$ images over $42$ steps with CFG scale $4.5$ and batch size 1. Each method runs in its own process and is averaged over ten runs after ten warm-up runs. In the standard deployment the register shares the frozen generator backbone and text encoders of the sampling pipeline, so the memory overhead is limited to the $147.6$\,M trainable parameters of the read path and reward head: peak memory rises only from $16.9$\,GiB to $17.6$\,GiB versus unguided CFG (a $1.04\times$ factor). Baselines that require separate neural backbones instead consume $27$--$34$\,GiB. Computationally, each guided step adds roughly $50$\,ms and $5.9$\,TFLOPs on top of the $12.4$\,TFLOPs of a base generation step.

End-to-end latency is $1.80\times$ that of standard CFG, slightly above PAVRM ($1.67\times$) but far below Diffusion Probe ($2.63\times$) and the decode-then-ImageReward pipeline ($4.12\times$). Restricting guidance to the early phase ($\tau_i>0.8$, i.e., $18$ of the $42$ steps) lowers the overhead to $1.34\times$ CFG while keeping roughly $80\%$ of the HPSv3 gains of the full scheduler.

\subsection{RGS across Sampling Budgets and Strengths}
\label{sec:appendix-step-scale}

\begin{figure}[t]
    \centering
    \includegraphics[width=1\linewidth]{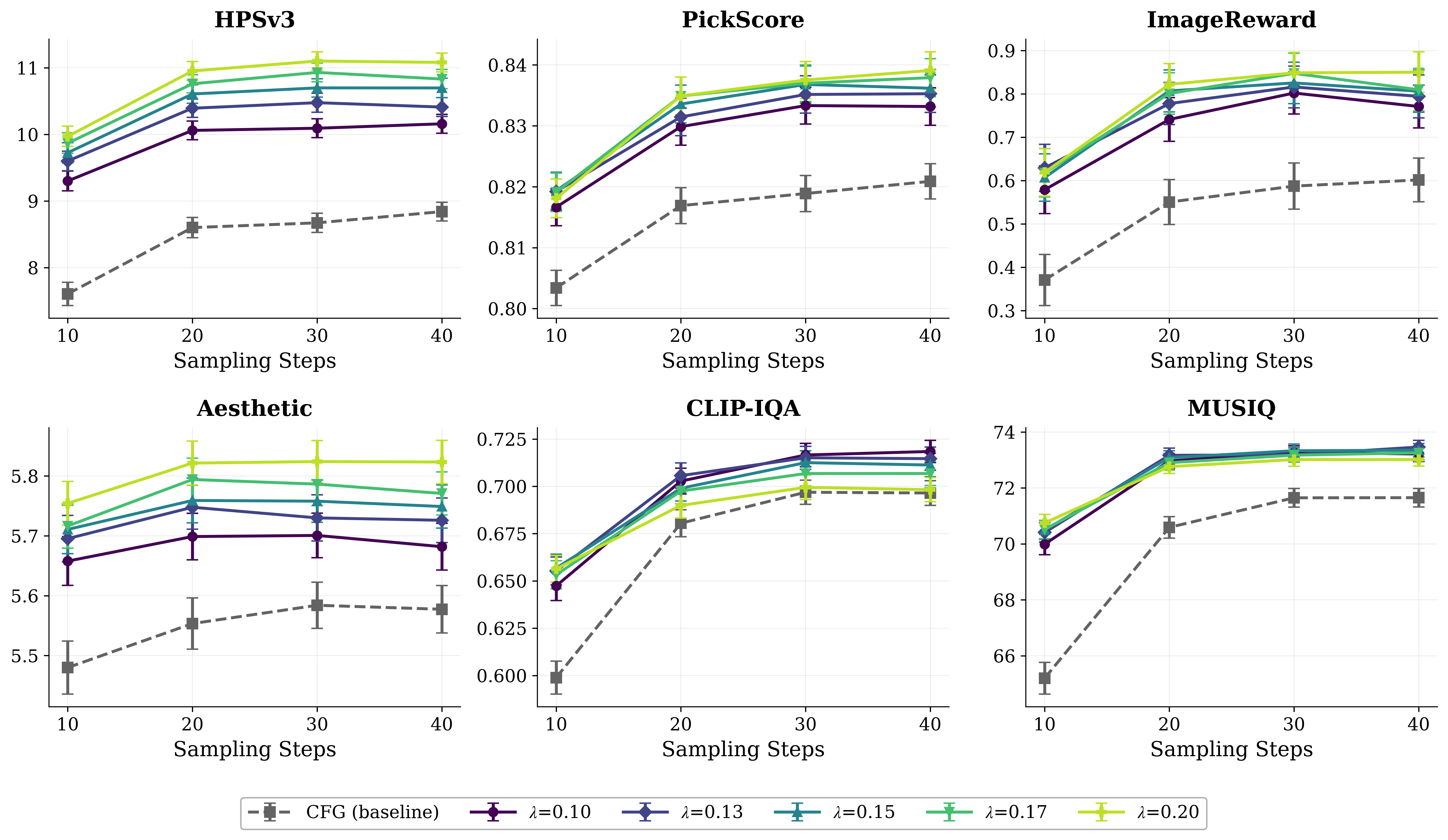}
    \caption{\textbf{Effect of sampling budgets and guidance strengths.} Each panel plots one metric against the total number of generation steps, comparing the CFG baseline with RGS under uniform guidance strengths $\guidance$ (labeled as $\lambda$). Target rewards increase with guidance strength, while no-reference perceptual quality remains above the CFG baseline and plateaus at higher strengths.}
    \label{fig:step-scale}
\end{figure}

Figure~\ref{fig:step-scale} evaluates uniform guidance strengths under sampling budgets of $10$, $20$, $30$, and $40$ steps, in contrast to the three-band scheduler used in the main deployment ($42$ steps). Guidance improves every metric relative to the CFG baseline at all budgets, and the targeted reward metrics generally rise with guidance strength. No-reference metrics such as CLIP-IQA and MUSIQ also exceed the CFG baseline but plateau, or vary only marginally, at higher strengths. The parameter $\lambda$ in the figure legends is mathematically equivalent to the guidance strength $\guidance$ defined in the method section.

\subsection{Reward along the RGS Trajectory}
\label{sec:appendix-reward-trajectory}

\begin{figure}[t]
    \centering
    \includegraphics[width=\linewidth]{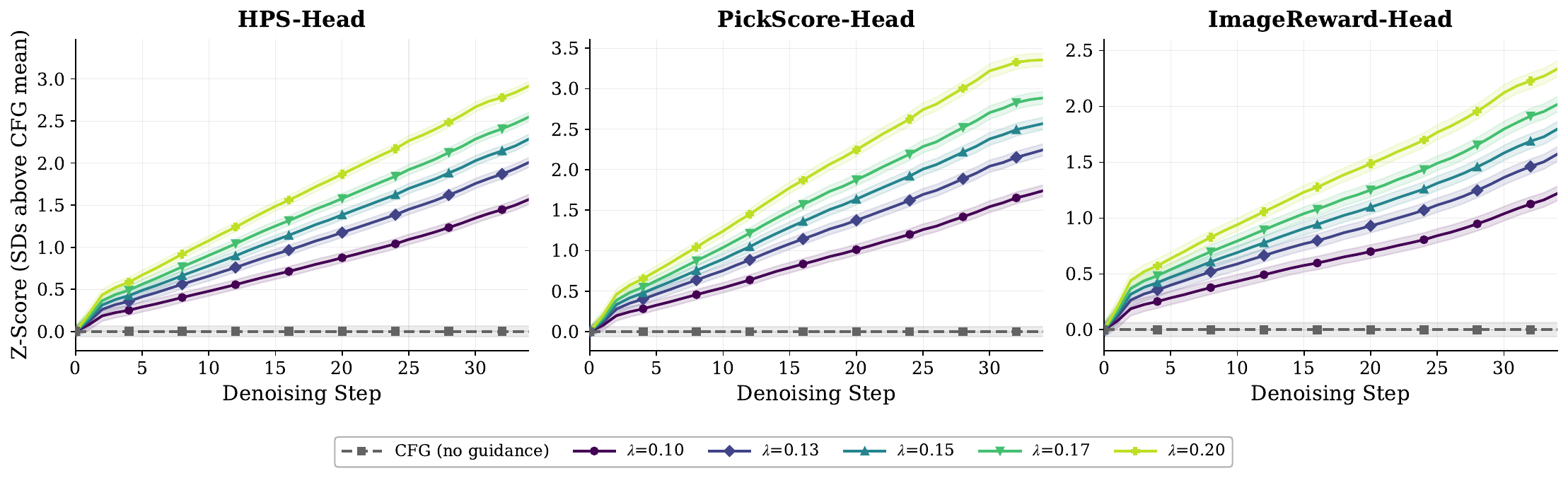}
    \caption{\textbf{Reward progression during generation under uniform guidance.} Register-head evaluations are plotted as $z$-scores, normalized against the per-step distribution of the unguided CFG baseline. Curves denote different guidance strengths $\guidance$ (labeled as $\lambda$ in the legend), with shaded regions indicating variance across diverse text prompts.}
    \label{fig:reward-trajectory}
\end{figure}

\textbf{The reward-gradient correction exhibits progressive accumulation along the generation trajectory, proportionally scaling with the specified guidance magnitude.}
This auxiliary analysis uses a uniform guidance strength $\guidance\in[0.10,0.20]$ rather than the three-band scheduler of the main deployment. Figure~\ref{fig:reward-trajectory} tracks the register-head scores for HPSv3, PickScore, and ImageReward through the denoising process. The correction raises the scores by $2$--$3.3$ standard deviations relative to the CFG baseline mean, and the ordering of guidance strengths is preserved across all three evaluators.

\subsection{Robustness to the ODE Solver}
\label{sec:appendix-ode-solver}

\begin{table}[t]
\centering
\small
\renewcommand{\arraystretch}{1.15}
\setlength{\tabcolsep}{3.5pt}

\caption{\textbf{Performance of RGS across ODE integrators.} For the CFG rows, FID$\downarrow$ is computed relative to the unguided Euler CFG reference set (integrator-induced drift); for the RGS rows, relative to the \emph{corresponding same-solver} CFG reference set (drift attributable to RGS). NFE denotes the number of function evaluations per step, $\guidance_{\mathrm{mid}}$ the mid-band guidance strength, and IR ImageReward. RGS improves target reward and perceptual quality metrics under every solver.}
\label{tab:ode-solver}

\resizebox{\linewidth}{!}{%
\begin{tabular}{@{}ll cccccc@{}}
\toprule
\textbf{Solver} & \textbf{Method} & \multicolumn{3}{c}{\textbf{Reward Model} $\uparrow$} & \multicolumn{3}{c}{\textbf{Quality / Mode drift}} \\
\cmidrule(lr){3-5}\cmidrule(l){6-8}
{\scriptsize(NFE)} & {\scriptsize$(\guidance_{\mathrm{mid}})$} & \textbf{HPSv3} & \textbf{IR} & \textbf{PickScore} & \textbf{FID}$\downarrow$ & \textbf{MUSIQ}$\uparrow$ & \textbf{CLIP-IQA}$\uparrow$ \\
\midrule
Euler {\scriptsize(1)} & CFG & 8.33 & 0.977 & 0.870 & 0.0 & 73.0 & 0.697 \\
 & +\method\ $(0.05)$ & 10.34 & 1.167 & 0.888 & 64.8 & 74.9 & 0.730 \\
 & +\method\ $(0.10)$ & 10.44 & 1.154 & 0.888 & 69.3 & 74.2 & 0.712 \\
\midrule
AB2 {\scriptsize(1)} & CFG & 8.29 & 0.995 & 0.869 & 40.1 & 72.8 & 0.686 \\
 & +\method\ $(0.05)$ & 10.31 & 1.167 & 0.887 & 66.5 & 74.8 & 0.731 \\
 & +\method\ $(0.10)$ & 10.37 & 1.160 & 0.887 & 72.2 & 74.2 & 0.716 \\
\midrule
Midpoint {\scriptsize(2)} & CFG & 8.35 & 0.996 & 0.870 & 45.0 & 72.7 & 0.681 \\
 & +\method\ $(0.05)$ & 10.14 & 1.167 & 0.887 & 63.4 & 74.7 & 0.728 \\
 & +\method\ $(0.10)$ & 10.25 & 1.155 & 0.886 & 69.1 & 74.2 & 0.715 \\
\midrule
Heun {\scriptsize(2)} & CFG & 8.27 & 1.008 & 0.869 & 42.6 & 72.8 & 0.687 \\
 & +\method\ $(0.05)$ & 10.11 & 1.167 & 0.887 & 64.2 & 74.6 & 0.727 \\
 & +\method\ $(0.10)$ & 10.25 & 1.155 & 0.886 & 69.7 & 74.1 & 0.716 \\
\midrule
RK4 {\scriptsize(4)} & CFG & 8.38 & 1.006 & 0.871 & 46.8 & 72.9 & 0.685 \\
 & +\method\ $(0.05)$ & 10.17 & 1.170 & 0.888 & 63.8 & 74.6 & 0.728 \\
 & +\method\ $(0.10)$ & 10.27 & 1.152 & 0.887 & 70.0 & 74.0 & 0.715 \\
\bottomrule
\end{tabular}%
} 
\end{table}

The primary results in the main text are based on first-order Euler integration of the generative flow ODE~\citep{lipman2023flow}. To verify whether RGS generalizes beyond this specific discretization, we evaluate the RGS protocol detailed in Appendix~\ref{sec:appendix-settings-guidance} using four additional classical numerical integrators~\citep{hairer1993solving}. These include the two-step Adams--Bashforth (AB2) method, which is a multistep method requiring a single model evaluation per step analogous to Euler; the explicit midpoint method; Heun's method (a trapezoidal predictor--corrector); and the classical fourth-order Runge--Kutta (RK4) method. RGS composes with higher-order solvers in the same way as with Euler: the reward gradient is computed once at the beginning of each step, the solver's internal velocity evaluations use standard CFG forwards, and the reward-gradient correction from Sec.~\ref{sec:reward-guided-sampling} is applied to the solver output afterwards. Its magnitude is scaled relative to the step displacement, $\guidance_i\cdot\rms(\Delta_{\mathrm{ref},i})$, with no hyperparameter re-tuning. For these experiments, we implement the three-band RGS schedule from Sec.~\ref{sec:reward-guided-sampling}. The guidance strengths are set to $\guidance_{\mathrm{early}}{=}0.30$ for the high-noise phase ($\tau_i{>}0.8$) and $\guidance_{\mathrm{mid}}\in\{0.05,0.10\}$ for the mid-range phase ($\tau_i{>}0.2$). The latter encompasses both the standard deployed configuration and an augmented mid-band variant, while guidance is omitted in the low-noise tail. The evaluation is conducted on a subset of $250$ prompts from the balanced benchmark, with two seeds per prompt. Generations are performed at $1024^2$ resolution over $42$ steps with a CFG scale of $4.5$ using single-head RGS. Each solver's respective CFG baseline and the two RGS configurations are evaluated on matched prompt--seed pairs. Because these runs use a $250$-prompt subset, the baseline CFG numbers deviate slightly from those in Table~\ref{tab:guided-comparison}, which uses the full $400$-prompt set.

The findings are summarized as follows:

\textbf{(1) RGS is robust across solvers.} As Table~\ref{tab:ode-solver} shows, RGS reproduces the gains seen with Euler under every integrator: paired per-image HPSv3 rises by roughly two points over the corresponding same-solver CFG baseline, with consistent improvements in the other optimized reward metrics and in no-reference quality. The drift attributable to RGS stays controlled relative to each solver's baseline, so the magnitude-matched correction generalizes across integrators without per-solver tuning or any compromise in no-reference quality.

\textbf{(2) Cross-solver FID mostly reflects trajectory decorrelation, not generative degradation.} Changing the integrator introduces an FID of $40$--$47$ relative to the Euler CFG baseline while the other six metrics stay similar. For a fixed latent seed, different solvers converge to distinct but qualitatively comparable endpoints, so the drift induced by guidance must be measured against the corresponding solver's own baseline, as in Table~\ref{tab:ode-solver}.

\textbf{(3) Higher-order solvers reduce the reward gain only marginally.} Midpoint, Heun, and RK4 yield HPSv3 gains of $+1.79$ to $+1.98$, versus $+2.01$ to $+2.11$ for Euler and AB2. We hypothesize that applying the reward displacement at the end of each step exposes it to the internal evaluations of higher-order solvers, which partially undo the off-manifold correction during the next integration step. The reduction is only $5$--$10\%$ of the total reward gain and does not change the qualitative conclusions. Since the single-evaluation AB2 solver closely matches Euler, both offer the best reward--quality trade-off among the integrators tested.

\subsection{Qualitative Comparison Across Prompts}
\label{sec:gallery}

\begin{figure}[t]
    \centering
    \includegraphics[width=0.95\linewidth]{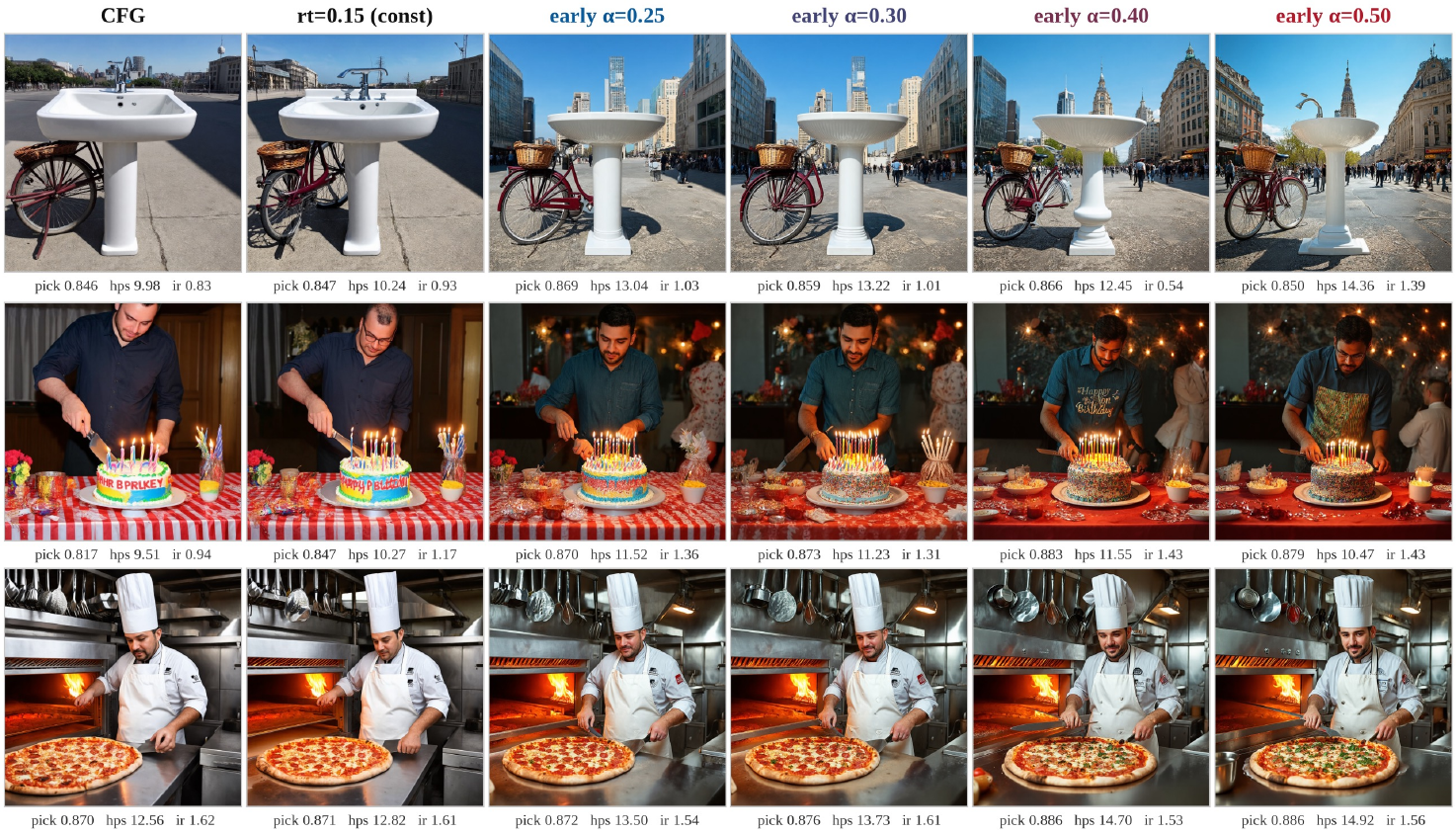}
    \caption{\textbf{Effect of guidance strength on generation.} Samples generated under monotonically increasing guidance strengths $\guidance$ with a fixed prompt and initial latent. Increasing $\guidance$ progressively strengthens preference-aligned details.}
    \label{fig:appendix-rt-scale}
\end{figure}

We additionally test RGS under a flat strength $\guidance{=}0.15$, complementing the deployed schedule in Table~\ref{tab:sampling-config}. Figure~\ref{fig:appendix-rt-scale} uses a matched prompt and seed to isolate the effect of guidance strength. Increasing $\guidance$ progressively instantiates stronger preference-aligned details.

\section{Future Work}
\label{sec:appendix-future}
Our results show that reward-gradient guidance applied to noisy latents under a frozen generator improves preference alignment while keeping no-reference quality intact. Two directions seem especially promising.

\textbf{(1) Few-Step Architectures and Video Generation.} A natural next step is to test whether latent reward registers transfer to few-step distilled samplers and video diffusion models. For video, the reward mechanism will need structural adaptations to enforce temporal consistency and coherent motion.

\textbf{(2) From RLHF to Reinforcement Learning from Verifiable Rewards (RLVR).} The framework is not limited to human-preference optimization. We plan to explore its integration with RLVR, where the same latent readout could optimize objective criteria such as factual fidelity, spatial reasoning, and physical plausibility.

\end{document}